\documentclass[letterpaper, 12pt]{article}

\usepackage[margin=1.375in]{geometry}

\usepackage{siunitx} % SI units and styles
\usepackage{url}
\usepackage{mathtools}
\usepackage{amsfonts, amssymb}
\usepackage{xfrac}
\usepackage{algpseudocode}
\usepackage{algorithm}
\usepackage{authblk}
\usepackage{graphicx}
\DeclareGraphicsExtensions{.pdf,.png,.jpg,.mps,.eps,.ps}
\usepackage{xcolor, soul}
\usepackage{multirow}
\usepackage{caption}
\usepackage[comma,numbers,sort&compress]{natbib}

\usepackage[font=scriptsize]{subcaption}
\usepackage[colorinlistoftodos]{todonotes}
\graphicspath{{figure/}}

\newcommand{\bm}[1]{\boldsymbol{\mathbf{#1}}}
\newcommand{\given}{\,|\,}
\newcommand{\tp}{{\top}}

\DeclareMathOperator{\argmax}{arg\max}
\DeclareMathOperator{\tr}{tr}

\DeclareMathOperator*{\cov}{cov}

\DeclareMathOperator{\diag}{diag}

\allowdisplaybreaks

\title{Variational Latent Gaussian Process for Recovering Single-Trial Dynamics from Population Spike Trains}
\author[1,2]{Yuan Zhao\thanks{yuan.zhao@stonybrook.edu}}
\author[1,2,3]{Il Memming Park\thanks{memming.park@stonybrook.edu}}
\affil[1]{Department of Neurobiology and Behavior}
\affil[2]{Department of Applied Mathematics and Statistics}
\affil[3]{Institute for Advanced Computational Sciences}
\affil[ ]{Stony Brook University, Stony Brook, NY, USA}
\date{}

\begin{document}
\maketitle

\begin{abstract}
When governed by underlying low-dimensional dynamics, the interdependence of simultaneously recorded population of neurons can be explained by a small number of shared factors, or a low-dimensional trajectory.
Recovering these latent trajectories, particularly from single-trial population recordings, may help us understand the dynamics that drive neural computation.
However, due to the biophysical constraints and noise in the spike trains, inferring trajectories from data is a challenging statistical problem in general.
Here, we propose a practical and efficient inference method, called the variational latent Gaussian process (vLGP).
The vLGP combines a generative model with a history-dependent point process observation together with a smoothness prior on the latent trajectories.
The vLGP improves upon earlier methods for recovering latent trajectories, which assume either observation models inappropriate for point processes or linear dynamics.
We compare and validate vLGP on both simulated datasets and population recordings from the primary visual cortex.
In the V1 dataset, we find that vLGP achieves substantially higher performance than previous methods for predicting omitted spike trains, as well as capturing both the toroidal topology of visual stimuli space, and the noise-correlation.
These results show that vLGP is a robust method with a potential to reveal hidden neural dynamics from large-scale neural recordings.
\end{abstract}

\section{Introduction}
Neural populations implement dynamics that produce robust behavior; however, our current experimental observations of these dynamics are invariably indirect and partial.
In classical analyses of neural spike trains, noisy responses are averaged over repeated trials that are presumably time-locked to a stereotypical computation process.
However, neural dynamics are not necessarily time-locked nor precisely repeated from trial to trial; rather, many cognitive processes generate observable variations in the internal processes that sometimes manifest in behavior such as error trials, broad reaction time distributions, and change of mind~\cite{Latimer2015,Jazayeri2010,Resulaj2009}.
In addition, it is difficult to disambiguate different possible neural implementations of computation from the average trajectory since they may only differ in their trial-to-trial variability~\cite{Latimer2015,Churchland2011}.
Therefore, if we wish to understand how neural computation is implemented in neural populations, it is imperative that we recover these hidden dynamics from individual trials~\cite{Paninski2009,Kao2015b}.

Advances in techniques for recording from larger subpopulations facilitate single-trial analysis, especially the inference of single-trial latent dynamical trajectories.
Several statistical approaches have been developed for extracting latent trajectories that describe the activity observed populations~\cite{Yu2009,Koyama2010,Macke2011c,Pfau2013,Archer2014f,Frigola2014}.
For example, latent trajectories recovered from motor cortex suggest that these methods can provide insight to the coding and preparation of planned reaching behavior~\cite{Sadtler2014,Churchland2012,Churchland2010}.
Latent trajectories also elucidate the low-dimensional noise structure of neural codes and computations~\cite{Moreno-Bote2014,Haefner2013,Sadtler2014,Ecker2014b}.

Inference of latent dynamical trajectories is a dimensionality-reduction method for multi-variate time series, akin to Kalman smoothing or factor analysis~\cite{Kao2015b}.
Given a high-dimensional observation sequence, we aim to infer a shared, low-dimensional latent process that explains the much of the variation in high-dimensional observations.
A large class of methods assume an autoregressive linear dynamics model in the latent process due to its computational tractability~\cite{Paninski2009,Macke2011c,Buesing2012,Kao2015b}, we refer to these as PLDS (Poisson Linear Dynamical System).
Although the assumption of linear dynamics can help in smoothing, it can also be overly simplistic: interesting neural computations are naturally implemented as nonlinear dynamics, and evidence points to nonlinear dynamics in the brain in general.
Therefore, we propose to relax this modeling assumption and
%Therefore, to understand how the brain implements computation, and how task-irrelevant noise is structured, we need to relax our modeling assumptions about the latent variables.
impose a general Gaussian process prior to nonparametrically infer the latent dynamics, similar to the Gaussian process factor analysis (GPFA) method~\cite{Yu2009,Lakshmanan2015}.
However, we differ from GPFA in that we use a point process observation model with self-history dependence rather than an instantaneous Gaussian observation model. 
A Gaussian observation model is inappropriate for inference in the millisecond-range time scale since it cannot generate spike counts.
The price we pay is a non-conjugate prior and, consequently, an approximate posterior inference~\cite{Paninski2009}.
We use a variational approximation~\cite{Blei2016} where we assume a Gaussian process posterior over the latents, and optimize a lower bound of the marginal likelihood for the inference.
Our algorithm, we call \textit{variational latent Gaussian process} (vLGP), is fast and has better predictability compared to both GPFA and PLDS
% as the reviewer asked 
at a fine timescale (1~ms bins).
We compare these algorithms on simulated systems with known latent processes.
We apply it to high-dimensional V1 data from anesthetized monkey to recover both the noise correlation structure and topological structure of population encoding of drifting orientation grating stimuli.

\section{Generative model}
\begin{figure}[t]
	\centering
	\includegraphics[width=\linewidth]{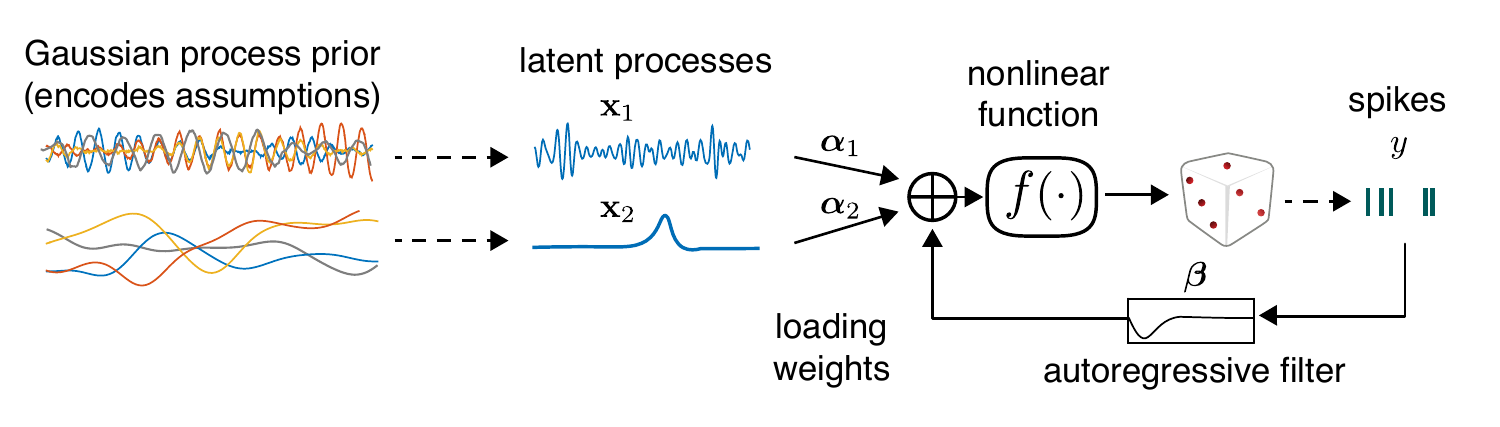}
	\caption{
		Generative model schematic representing~\eqref{eq:intensity} for one neuron driven by two latent processes.
		Every neuron in the observed population are driven by the same set of latent processes.
		The inferred latent processes are more likely to be smooth, as assumed by the smooth Gaussian process prior.
		Given $\bm{x}_t$ and $\bm{\alpha}_n$, the spike train $y$ is generated as a generalized linear model (GLM)~\cite{Truccolo05,Pillow08}.
		The point nonlinearity is fixed to be exponential $f(\cdot) = \exp(\cdot)$.
	}
	\label{fig:schematic}
\end{figure}

Suppose we simultaneously observe spike trains from $N$ neurons.
Let $\left(y_{t,n}\right)_{t=1,\ldots,T}$ denote the spike count time-series from the $n$-th neuron for a small time bin.
We model noisy neural spike trains mathematically as a simple point process which is fully described by its conditional intensity function~\cite{Daley1988}.
We assume the following parametric form of the conditional intensity function $\lambda^\ast(\cdot)$ for the point process log-likelihood~\cite{Daley1988,Macke2011c}:
\begin{equation}\label{eq:intensity}
\begin{split}
\log p(y_{t, n} \given \bm{x}_t, \bm{h}_{t,n}, \bm{\alpha}_n, \bm{\beta}_n) &=
y_{t,n} \log \lambda^\ast(t,n \given \bm{h}_{t,n}) - \lambda^\ast(t,n \given \bm{h}_{t,n}),
\\
\lambda^\ast(t,n \given \bm{h}_{t,n}) &= \exp\left(
\bm{\alpha}_n^\tp \bm{x}_t + \bm{\beta}_n^\tp \bm{h}_{t,n}
\right),
\end{split} 
\end{equation}
where $\bm{x}_t$ is a latent process and $\bm{h}_{t,n} = \left[ 1, y_{t-p,n}, y_{t-p+1,n}, \ldots, y_{t-1,n} \right]^\tp$ denotes the spike history vector~\cite{Truccolo05,Pillow08}.
Each neuron is directly influenced by the observed self-history\footnote{It is straightforward to add external covariates similar to the self-history in this point process regression~(e.g., see \citet{Park2014d}).} with weight $\bm{\beta}_n$ and also driven by the common latent process with weight $\bm{\alpha}_n$ (Fig.~\ref{fig:schematic}).
Neurons are conditionally independent otherwise: all trial-to-trial variability is attributed either to the latent process or individual point process noise (c.f., ~\citet{Goris2014,Ecker2014b,Lin2015}).

The vector $\bm{x}_t$ denotes the $L$-dimensional latent process at time $t$.
We assume that $L \ll N$, since we are looking for a small number of latent processes that explain the structure of a large number of observed neurons.
The vector $\bm{\beta}_n$ consists of the weights of the spike history and a time-independent bias term of the log firing rate for each neuron, and $\bm{h}_{t,n}$ is a vector of length $(1 + p)$ containing the dummy value $1$ for the bias and $p$ time-step spike self-history.
This parametrization assumes that at most $p$ bins in the past influence the current intensity.

Under conditional independence, the joint distribution (data likelihood) of $N$ spike trains is given by,
\begin{equation}
p(y_{1 \ldots T, 1\ldots N} \given \bm{x}_{1 \ldots T}, \bm{\alpha}_{1 \ldots N}, \bm{\beta}_{1 \ldots N}) 
= \prod_{t=1}^T \prod_{n=1}^N p(y_{t, n} \given \bm{x}_t, \bm{h}_{t,n}, \bm{\alpha}_n, \bm{\beta}_n).
\end{equation}

Note that this model is not identifiable since
$\bm{\alpha}_n^T \bm{x}_t = 
(\bm{\alpha}_n^T \bm{C}) (\bm{C}^{-1} \bm{x}_t) = 
\bm{\alpha'}_n^T \bm{x'}_t$
where $\bm{C}$ is an arbitrary $L \times L$ invertible matrix
(see later sections for further discussions).
Also, the mean of latent process $\bm{x}$ can be traded off with the bias term in $\bm{\beta}$.

Our assumptions about the latent process---namely the smoothness over time in this paper---are encoded in the prior distribution over the latent process.
We use the Gaussian process (GP) framework~\cite{Rasmussen2005} for flexible prior design of each dimension $x_l(t)$ independently:
\begin{align}\label{eq:gp-general}
x_l(t) &\sim \mathcal{GP}(\mu_l, \kappa_l)
\end{align}
where $\mu_l(t)$, and $\kappa_l(t,s)$ are mean and covariance functions, respectively.
When time is discretized, the GP prior reduces to a multi-variate Gaussian distribution over the latent time series.
We use the following form:
\begin{equation}
p(\bm{x}_l) =
\mathcal{N}(\bm{x}_l \given \bm{0}, \bm{K}_l), \qquad l = 1, \ldots, L.
\end{equation}
For the analyses in this manuscript, we choose the squared exponential covariance function~\cite{Rasmussen2005} for general smoothness over time,
\begin{equation}
\cov(x_{t,l}, x_{s,l}) = \sigma_l^2\exp(-\omega_l(t-s)^2).
\end{equation}
where $\sigma_l$ and $\omega_l$ are hyperparameters corresponding to the magnitude and inverse time scale of the latent process, respectively.

\section{Variational inference}
Our goal is to infer the posterior distribution over the latent process and fit the model parameters given the observed data.
By Bayes' theorem, the posterior distribution of the latent process is,
\begin{equation}
p(\bm{x}_{1 \ldots L} \given \bm{y}_{1 \ldots N}) = \frac{p(\bm{y}_{1 \ldots N} \given \bm{x}_{1 \ldots L}) p(\bm{x}_{1 \ldots L})}{p(\bm{y}_{1 \ldots N})},
\end{equation}
However, unlike in GPFA, the posterior under a point process likelihood and Gaussian process prior does not have an analytical form~\cite{Paninski2009}.
Consequently, we must turn to an approximate inference technique. 
We employ variational inference, which aims to find an approximate distribution $q(\bm{x})$ of the intractable true posterior $p(\bm{x} \given \bm{y})$.
We can introduce this approximate posterior into the likelihood by re-writing it as,
\begin{align}
\log p(\bm{y}_{1 \ldots N}) &= \mathcal{E}_q[\log p(\bm{y}_{1 \ldots N})] 
= \mathcal{E}_q\left[\log \frac{p(\bm{y}_{1 \ldots N}, \bm{x}_{1 \ldots L})}{q(\bm{x}_{1 \ldots L})} \cdot \frac{q(\bm{x}_{1 \ldots L})}{p(\bm{x}_{1 \ldots L} \given \bm{y}_{1 \ldots N})}\right] \\
&= \underbrace{\mathcal{E}_q \left[\log \frac{p(\bm{y}_{1 \ldots N}, \bm{x}_{1 \ldots L})}{q(\bm{x}_{1 \ldots L})}\right]}_{\mathcal{L}(q)}
+ 
\underbrace{\mathcal{E}_q\left[\log \frac{q(\bm{x}_{1 \ldots L})}{p(\bm{x}_{1 \ldots L} \given \bm{y}_{1 \ldots N})}\right]}_{D_{\mathrm{KL}}(q \| p)},
\end{align}
where $\mathcal{E}_q$ denotes an expectation over $q(\bm{x})$, and $D_{\mathrm{KL}}(q \| p)$ is the Kullback-Leibler divergence, which measures the difference in the true posterior and its variational approximation.
Since $D_{\mathrm{KL}}(q \| p)$ is non-negative, $\mathcal{L}(q)$ is the lower bound for the marginal likelihood.
Finding an approximate posterior $q$ close to the true posterior by minimizing the Kullback-Leibler divergence is equivalent to maximizing the lower bound $\mathcal{L}(q)$, also known as the Evidence Lower BOund (ELBO).

We further assume that the $q$ distribution factorizes into Gaussian distributions with respect to each dimension of the latent process, such that
\begin{equation}
q(\bm{x}_{1 \ldots L}) = \prod_{l=1}^L \mathcal{N}(\bm{x}_l \given \bm{\mu}_l, \bm{\Sigma}_l).
\end{equation}
We then obtain the lower bound:
\begin{equation}\label{eq:ELBO}
\begin{split}
\mathcal{L}(q) =& \sum_{t=1}^T \sum_{n=1}^N \mathcal{E}_q[\log p(y_{t, n} \given \bm{x}_t, \bm{h}_{t,n}, \bm{\alpha}_n, \bm{\beta}_n)] - \sum_{l=1}^L \mathcal{E}_q\left[\log \frac{q(\bm{x}_{1 \ldots L} \given \bm{\mu}_l, \bm{\Sigma}_l)}{p(\bm{x}_{1 \ldots L} \given \bm{K}_l)}\right]
\\
=& \sum_{t=1}^T \sum_{n=1}^N [y_{t, n}(\bm{\alpha}_n^\tp \bm{\mu}_t + \bm{\beta}_n^\tp \bm{h}_{t,n}) - \exp(\bm{\alpha}_n^\tp \bm{\mu}_t + \bm{\beta}_n^\tp \bm{h}_{t,n} + \frac{1}{2} \bm{\alpha}_n^\tp \bm{\Sigma}_t \bm{\alpha}_n)]
\\
& - \frac{1}{2} \sum_{l=1}^L [\bm{\mu}_l^\tp \bm{K}_l^{-1} \bm{\mu}_l + \tr(\bm{K}_l^{-1} \bm{\Sigma}_l) - \log\det(\bm{K}_l^{-1}\bm{\Sigma}_l) - T].
\end{split}
\end{equation}
where $T$ is the number of total time steps, and each temporal slice $\bm{\mu}_t$ is a vector of posterior means of the $L$ latent variables at time $t$.
Each temporal slice $\bm{\Sigma}_t$ is a diagonal matrix whose diagonal contains the variances of the $L$ latent variables at time $t$.

Variational inference for the entire posterior over latents, parameters, and hyperparameters can all be formulated in terms of maximizing~\eqref{eq:ELBO}.
We sequentially update all parameters coordinate-wise; each conditional update turns out to be a convex-optimization problem except for the hyperparameters as explained below.
We derive the inference algorithm (vLGP) in the following sections, and it is summarized in Algorithm \ref{pseudocode}.

Our algorithm scales linearly in space $\mathcal{O}(Ts)$ and time $\mathcal{O}(Tr^2L)$ per iteration (for a fixed hyperparameter) where $s = \max(rL, pN)$ thanks to the rank-$r$ incomplete Cholesky factorization of the prior covariance matrix.
For comparison, time complexity of GPFA is $\mathcal{O}(T^3L^3)$, and that of PLDS is $\mathcal{O}(T(L^3 + LN))$.

\begin{algorithm}[p]
	\caption{Pseudocode for vLGP inference}\label{pseudocode}
	\begin{algorithmic}[1]
		\fontsize{12}{10}\selectfont
		\Procedure{vLGP}{$\bm{y}_{1 \ldots T}$, $\bm{h}_{1 \ldots T, 1 \ldots N}$, $\sigma_{1 \ldots L}^2$, $\omega_{1 \ldots L}, tol, k$}
		
		\State $\bm{G}_l = \mathrm{ichol}(\sigma_l^2, \omega_l),\, l=1 \ldots L$ \Comment{construct incomplete Cholesky decomposition~\cite{Bach2002}}
		\State Initialize $\bm{\alpha}_n$ and $\bm{\mu}_l$ by factor analysis
		\State $\bm{\beta}_n \gets (\bm{h}_{1 \ldots T, n}^\tp\bm{h}_{1 \ldots T, n})^{-1} \bm{h}_{1 \ldots T, n}^\tp \bm{y}_{1 \ldots T},\, n=1 \ldots N$ \Comment{linear regression}
		
		\While{true}
		\For{$l \gets 1, \ldots, L$}
		\State $\lambda_{t, n} \gets \bm{\alpha}_n^\tp \bm{\mu}_t + \bm{\beta}_n^\tp \bm{h}_{t,n} + \frac{1}{2}\bm{\alpha}_n^\tp \bm{\Sigma}_t \bm{\alpha}_n,\, t = 1 \ldots T, n = 1 \ldots N$
		\State $\bm{u}_l \gets \bm{G}_l\bm{G}_l^\tp(\bm{y} - \bm{\lambda})\bm{\alpha}_l - \bm{\mu}_l^{old}$
		\State $\bm{B}_l \gets \bm{G}_l^\tp \diag(\bm{W}_{l}) \bm{G}_l$
		\State $\bm{\mu}_l^{new} \gets \bm{\mu}_l^{old} + [\bm{I}_T - \bm{G}_l \bm{G}_l^\tp \bm{W}_l + \bm{G}_l \bm{B}_l (\bm{I}_r + \bm{B}_l)^{-1} \bm{G}_l^\tp \bm{W}_l] \bm{u}_l$ \Comment{Newton-step for $\bm{\mu}$}
		\State $\bm{\mu}_l^{new} \gets (\bm{\mu}_l^{new} - \bar{\bm{\mu}}_l^{new})$ \Comment{constrain $\bm{\mu}$}
		\EndFor
		
		\For{$n \gets 1, \ldots, N$}
		\State $\lambda_{t, n} \gets \bm{\alpha}_n^\tp \bm{\mu}_t + \bm{\beta}_n^\tp \bm{h}_{t,n} + \frac{1}{2}\bm{\alpha}_n^\tp \bm{\Sigma}_t \bm{\alpha}_n,\, t = 1 \ldots T, n = 1 \ldots N$
		\State $\bm{\alpha}_n^{new} \gets \bm{\alpha}_n^{old} + [(\bm{\mu} + \bm{V} \circ \bm{\alpha}_n^{old})^\tp \mathrm{diag}(\bm{\lambda}_n) (\bm{\mu} + \bm{V} \circ \bm{\alpha}_n^{old}) + \mathrm{diag}(\bm{V}^\tp \bm{\lambda}_n)]^{-1} [\bm{\mu}^\tp \bm{y}_n - (\bm{\mu} + \bm{V} \circ \bm{\alpha}_n^{old})^\tp \bm{\lambda}_n]$ \Comment{Newton-step for $\bm{\alpha}$}
		\State $\bm{\beta}_n^{new} \gets \bm{\beta}_n^{old} + [\bm{h}_{n}^\tp \mathrm{diag}(\bm{\lambda}_n) \bm{h}_{n}]^{-1} \bm{h}_{n}^\tp (\bm{y}_n - \bm{\lambda}_n)$ \Comment{Newton-step for $\bm{\beta}$}
		\EndFor
		
		\State $\bm{\alpha}_l^{new} \gets \bm{\alpha}_l^{new} / \|\bm{\alpha}_l^{new}\|,\, l = 1 \ldots L$ \Comment{constrain $\bm{\alpha}$}	  
		
		\State $W \gets \bm{\lambda} \bm{\alpha}^{2\tp}$ \Comment{update diagonals of $W$}
		\State $\bm{B}_l \gets \bm{G}_l^\tp \diag(\bm{W}_l) \bm{G}_l,\, l = 1 \ldots L$
		\State $\bm{V}_{1 \ldots T, l} \gets [\bm{G}_l \circ (\bm{G}_l - \bm{G}_l \bm{B}_l + \bm{G}_l \bm{B}_l (I_k + \bm{B}_l)^{-1} \bm{B}_l)] \bm{1},\, l = 1 \ldots L$
		
		\State Optimize hyperparameters with the gradient in \eqref{eq:hypergrad} and update $\bm{G}_{1 \ldots L}$ every $k$ iterations
		\If{$\Vert(\bm{\mu}_{1 \ldots L}^{new}, \bm{\alpha}_{1 \ldots N}^{new}, \bm{\beta}_{1 \ldots N}^{new}) - (\bm{\mu}_{1 \ldots L}^{old}, \bm{\alpha}_{1 \ldots N}^{old}, \bm{\beta}_{1 \ldots N}^{old})\Vert < tol$}
		\State break
		\EndIf
		
		\State $
		\bm{\mu}_{1 \ldots L}^{old} \gets \bm{\mu}_{1 \ldots L}^{new},\,
		\bm{\alpha}_{1 \ldots N}^{old} \gets \bm{\alpha}_{1 \ldots N}^{new},\,
		\bm{\beta}_{1 \ldots N}^{old} \gets \bm{\beta}_{1 \ldots N}^{new}$
		
		\EndWhile
		\EndProcedure
	\end{algorithmic}
\end{algorithm}

\subsection{Posterior over the latent process}
The variational distribution $q_l$ is assumed to be Gaussian and thus determined only by its mean $\bm{\mu}_l$ and covariance $\bm{\Sigma}_l$.
The optimal solution is therefore obtained by
\begin{equation}
\bm{\mu}_{1 \ldots L}^\star, \bm{\Sigma}_{1 \ldots L}^\star = \underset{\bm{\mu}_{1 \ldots L}, \bm{\Sigma}_{1 \ldots L}}{\argmax} \mathcal{L}(q),
\end{equation}
while holding other parameters and hyperparameters fixed.

Denote the expected firing rate of neuron $n$ at time $t$ by $\lambda_{t, n}$,
\begin{equation}
\lambda_{t,n} 
= 
\mathcal{E}_q\left[ \lambda^\ast(t,n \given \bm{h}_{t,n}) \right]
=
\exp\left(
\bm{\beta}_n^\tp \bm{h}_{t,n} 
+ \bm{\alpha}_n^\tp \bm{\mu}_t 
+ \frac{1}{2} \bm{\alpha}_n^\tp \bm{\Sigma}_t \bm{\alpha}_n
\right).
\end{equation}
The optimal $\bm{\mu}_l$ can be obtained by the Newton-Raphson method.
The gradient and Hessian are given as
\begin{align}
\nabla_{\bm{\mu}_l} \mathcal{L}   =& \sum_{t,n} (y_{t,n} - \lambda_{t,n}) a_{n,l}\bm{e}_t - \bm{K}_l^{-1} \bm{\mu}_l,
\\
\nabla_{\bm{\mu}_l}^2 \mathcal{L} =& -\sum_{t,n} \lambda_{t,n} a_{n,l}^2 \bm{e}_t\bm{e}_t^\tp - \bm{K}_l^{-1}.
\end{align}
where $\bm{e}_t$ is a vector of length $T$ with value 1 at $t$ and zero elsewhere.
Note that the Hessian is negative definite, and hence this is a convex optimization given the other arguments and $\lambda_{t,n}$.
In each iteration, the update is
\begin{equation}
\bm{\mu}_l^{new} = \bm{\mu}_l^{old} 
- (\nabla_{\bm{\mu}_l}^2 \mathcal{L})^{-1}
(\nabla_{\bm{\mu}_l} \mathcal{L}).
\end{equation}

If we set the derivative w.r.t. $\bm{\Sigma}_l$ to 0,
\begin{equation}
\nabla_{\bm{\Sigma}_l} \mathcal{L} = -\frac{1}{2} \sum_{t,n} \lambda_{n,t} a_{n,l}^2 \bm{e}_t \bm{e}_t^\tp - \frac{1}{2} \bm{K}_l^{-1} + \frac{1}{2} \bm{\Sigma}_l^{-1} = 0,
\end{equation}
we obtain the optimal covariance,
\begin{align}
\bm{\Sigma}_l 
&= \left(\bm{K}_l^{-1} + \sum_{t,n} \lambda_{t,n} a_{n,l}^2 \bm{e}_t \bm{e}_t^\tp\right)^{-1}
\\
&= \left(\bm{K}_l^{-1} + \bm{W}_l\right)^{-1}.
\end{align}
where $\bm{W}_l = \sum_{t,n} \lambda_{t,n} a_{n,l}^2 \bm{e}_t \bm{e}_t^\tp$ is a diagonal matrix.
Therefore, there is no need for optimization of the covariance.
This simple form of variational posterior covariance has been noted before~\cite{Opper2009}.
Also note that $\nabla_{\bm{\mu}_l}^2 \mathcal{L} = -\bm{\Sigma}_l^{-1}$.

There is a redundancy between the bias term in $\bm{\beta}$ and the mean $\bm{\mu}$.
During optimization, we constrain the latent mean $\bm{\mu}$ by zero-centering, and normalize the loading $\bm{\alpha}$ by its max-norm latent-wise.

The prior covariance matrix $\bm{K}_l$ is large ($T \times T$) and is often severely ill-conditioned.
We only keep a truncated incomplete Cholesky factor $\bm{G}$~\cite{Bach2002} of size $T \times r$ where $r$ is the rank of the resulting approximation,
\begin{equation}
\bm{K}_l \approx \bm{G}_l \bm{G}_l^\tp,
\end{equation}
which provides both a compact representation and numerical stability.
Now, we derive key quantities that are necessary for a memory-efficient and numerically stable implementation.
For convenience and without ambiguity, we omit the subscript $l$ of all vectors and matrices below.
By the matrix inversion lemma~\cite{Rasmussen2005}, we have
\begin{equation}
\bm{\Sigma} = (\bm{K}^{-1} + \bm{W})^{-1} = \bm{K} - \bm{K} (\bm{W}^{-1} + \bm{K})^{-1} \bm{K}.
\end{equation}
and applying the lemma again
\begin{equation}
(\bm{W}^{-1} + \bm{K})^{-1} = \bm{W} - \bm{W} \bm{G} (\bm{I} + \bm{B})^{-1} \bm{G}^\tp \bm{W},
\end{equation}
where $\bm{B} = \bm{G}^\tp \bm{W} \bm{G}$.
We obtain two useful identities as a result:
\begin{align}
\bm{\Sigma}        &= \bm{G}\bm{G}^\tp - \bm{G} \bm{B} \bm{G}^\tp + \bm{G} \bm{B} (\bm{I} + \bm{B})^{-1} \bm{B} \bm{G}^\tp \label{eq:v},
\\
\bm{K}^{-1} \bm{\Sigma} &= \bm{I} - \bm{W} \bm{G}\bm{G}^\tp + \bm{W} \bm{G} (\bm{I}_k + \bm{B})^{-1} \bm{B} \bm{G}^\tp \label{eq:sv}.
\end{align}
With (\ref{eq:v}) and (\ref{eq:sv}), we can avoid large matrices in above equations such as,
\begin{align}
\tr[\bm{K}^{-1} \bm{\Sigma}]                   &= T - \tr[\bm{B}] + \tr[\bm{B} (\bm{I} + \bm{B})^{-1} \bm{B}],
\\
\log\det[\bm{K}^{-1} \bm{\Sigma}]               &= \log\det[\bm{I} - \bm{B} + \bm{B} (\bm{I} + \bm{B})^{-1} \bm{B}],
\\
\diag(\bm{\Sigma})                        &= [\bm{G} \circ (\bm{G} - \bm{G} \bm{B} + \bm{G} \bm{B} (\bm{I}_k + \bm{B})^{-1} \bm{B})] \bm{1},
\\
\bm{\Sigma} \nabla_{\bm{\mu}} \mathcal{L} &= (\bm{I} - \bm{G} \bm{G}^\tp \bm{W} + \bm{G} \bm{B} (\bm{I}_k + \bm{B})^{-1} \bm{G}^\tp \bm{W})\bm{u},
\end{align}
where $\bm{1}$ is the all-ones vector, and $\bm{u} = \bm{G}\bm{G}^\tp (\bm{y} - \bm{\lambda}) \bm{\alpha}_l - \bm{\mu}$. In addition, by the one-to-one correspondence between $\bm{W}$ and $\bm{\Sigma}$, we use the diagonal of $\bm{W}$ as a representation of $\bm{\Sigma}$ in the algorithm.

\subsection{Weights}
Denote the temporal slices of $\bm{\Sigma}_l$'s by $T \times L$ matrix $\bm{V}$. The optimal weights $\bm{\alpha}_n$ and $\bm{\beta}_n$ given the posterior over the latents can be obtained by the Newton-Raphson method with the following derivatives and Hessians,
\begin{align}
\nabla_{\bm{a}_n} \mathcal{L}   =& 
\bm{\mu}^\top (\bm{y}_n - \bm{\lambda}_n) - \diag(\bm{V}^\tp \bm{\lambda}_n) \bm{a}_n,
\\
\nabla_{\bm{a}_n}^2 \mathcal{L} =&
- (\bm{\mu} + \bm{V} \circ \bm{1}\bm{a}_n^\tp)^\tp \diag(\bm{\lambda}_n) (\bm{\mu} + \bm{V} \circ \bm{1}\bm{a}_n^\tp) - \diag(\bm{V}^\tp \bm{\lambda}_n),
\end{align}
and
\begin{align}
\nabla_{\bm{\beta}_n} \mathcal{L}   =& \bm{h}_{n}^\tp (\bm{y}_{n} - \bm{\lambda}_{n}),
\\
\nabla_{\bm{\beta}_n}^2 \mathcal{L} =& -\bm{h}_{n}^\tp \diag(\bm{\lambda}_{n}) \bm{h}_{n}.
\end{align}
The updating rules are
\begin{align}
\bm{\alpha}_n^{new} =& \bm{\alpha}_n^{old} - (\nabla_{\bm{\alpha}_n}^2 \mathcal{L})^{-1} \nabla_{\bm{\alpha}_n} \mathcal{L}, \label{eq:hypergrad}
\\
\bm{\beta}_n^{new}  =& \bm{\beta}_n^{old} - (\nabla_{\bm{\beta}_n}^2 \mathcal{L})^{-1} \nabla_{\bm{\beta}_n} \mathcal{L}.
\end{align}
Once again, both Hessians are negative definite, and hence in the territory of convex optimization.

\subsection{Hyperparameters}
One way to choose hyperparameters is to maximize the marginal likelihood w.r.t. the hyperparameters. Since the marginal likelihood is intractable in the vLGP model, we instead maximize \eqref{eq:ELBO} once again given the parameters and posterior. Interestingly, this objective function takes the same form as the one that is maximized in the GPFA's hyperparameters updates.

We write the squared-exponential covariance kernel as,
\begin{equation}
\bm{K}_l = \sigma_l^2 \exp(- \omega_l \bm{D}),
\end{equation}
where $\bm{D}$ is the matrix of squared distances of each time pair.
Hyperparameters $\sigma^2$ and $\omega$ corresponds to prior variance and inverse (squared) time scale.
We optimize the log-transformed hyperparameter for those are positive. To the $j$-th transformed hyperparameter of the $l$-th latent dimension, $\theta_{lj}$, the derivative is given as
\begin{align}
\frac{\partial \mathcal{L}}{\partial \theta_{lj}} &= \tr\left(\frac{\partial \mathcal{L}}{\partial \bm{K}_l} \frac{\partial \bm{K}_l}{\partial \theta_{lj}}\right),
\\
\frac{\partial \mathcal{L}}{\partial \bm{K}_l} &= \frac{1}{2}\left(\bm{K}_l^{-1} \bm{\mu}_l\bm{\mu}_l^\tp \bm{K}_l^{-1} + \bm{K}_l^{-1} \bm{\Sigma}_l \bm{K}_l^{-1} - \bm{K}_l^{-1}\right).
\end{align}
The optimal value can be found by common gradient algorithms for each latent dimension independently. 

The above derivation of the hyperparameter optimization technique assumes a fixed posterior and parameters. Thus it requires complete prior covariance matrices and explicit posterior covariance matrices rather than low-rank decompositions.
In order to avoid numerical singularity, we add a small quantity to the diagonal of prior covariance matrices.
It would be extremely costly to use these complete covariance matrices for long, consecutive time series.
Therefore, we randomly take many shorter temporal subsamples of the posterior for fast computation~\cite{Yu2009}.
One hyperparameter iteration is performed every fixed number of iterations of posterior and parameter optimization.
% We constrain the each hyperparameter to be updated at most by a factor of 5 on each iteration.

\section{Results}
We verified our inference algorithm recover the true parameters and latent variables when there is no model mismatch and then apply it to two simulated systems and one real dataset.
We compare our method (vLGP) against GPFA and PLDS.

%%%%%%%%%%%%%%%
% convergence %
%%%%%%%%%%%%%%%
\subsection{Convergence}
First of all, we demonstrate that vLGP converges to the correct parameters and latent variables under the assumed generative model.
We applied our method to simulated spike trains driven by 2-dimensional Gaussian process.
We fixed the number of time bins of a trial, GP variance and timescale ($T=200, \sigma^2=1, \omega = 0.01$). The values of parameters were randomly drawn from standard normal distribution. 
There are two limits that we need to consider for the convergence; increasing the duration of observations (more trials), and increasing the number of observed neurons.
The parameters and latent variables are initialized by factor analysis (FA). To identify the property of the global optima, we also initialize the parameters and latent variables at the values near the true ones (by adding zero mean and $0.1$ standard deviation Gaussian noises).

\begin{figure}[htb!]
	\centering
	\begin{subfigure}{\linewidth}
		\centering
		\includegraphics[width=\linewidth]{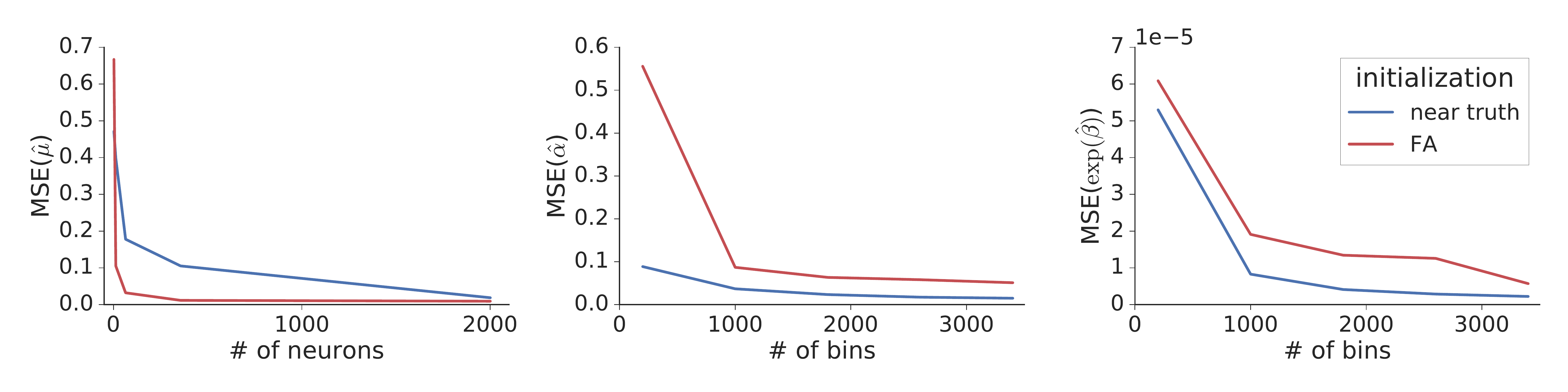}
		\caption{}\label{fig:converge}
	\end{subfigure}
	\begin{subfigure}{0.7\linewidth}
		\centering
		\includegraphics[width=\linewidth]{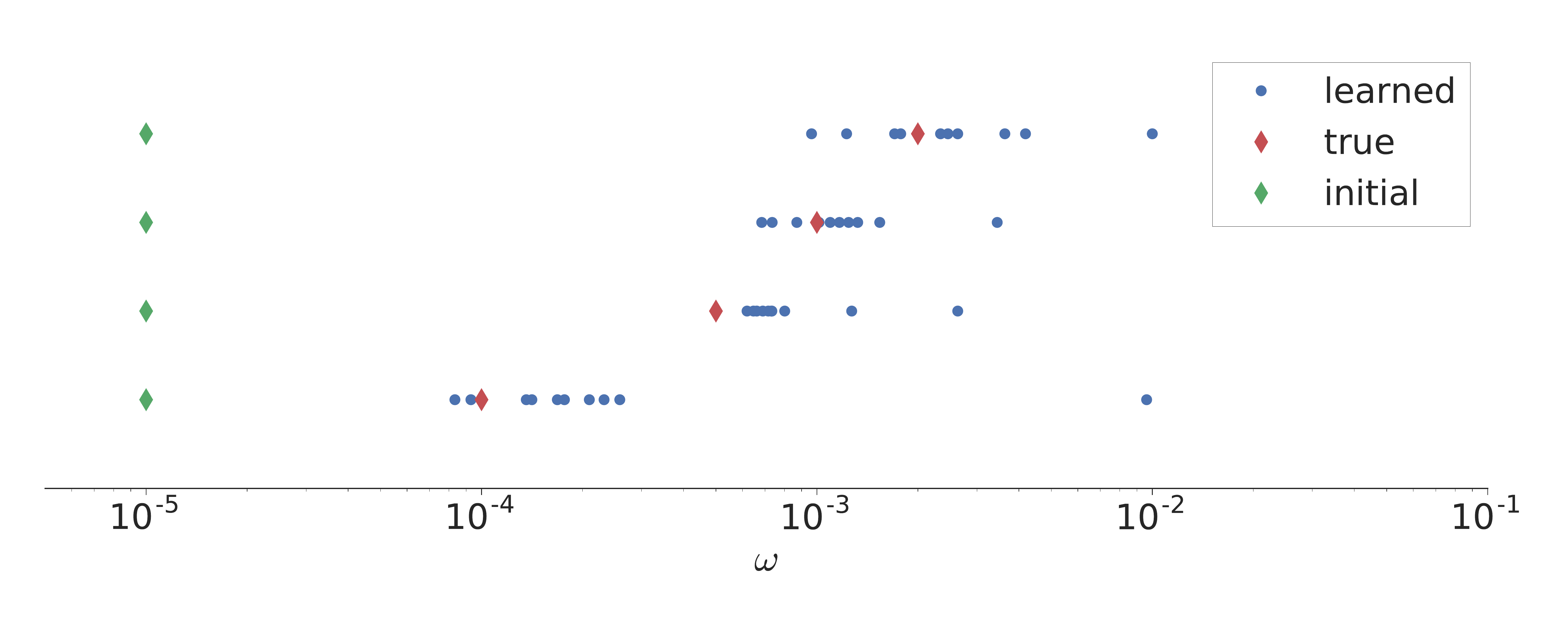}
		\caption{}\label{fig:hyper}
	\end{subfigure}
	\caption{vLGP learning under the assumed model. 
		(a) Convergence of latent variable and parameters vLGP. Notice that we use $\exp(\hat{\bm{\beta}})$ instead of the raw $\hat{\bm{\beta}}$ because a tiny deviation in the base firing rate results in a huge difference in the bias term $\hat{\beta}_0$ through the $\exp/\log$ transform.
		MSE was computed over a grid over number of neurons and trials.
		We plot the median MSE.
		(b) Learned timescales $\omega$. Green diamonds denote the initial values, red diamonds denote the true values, and blue dots denote learned values. For every true value, we simulated 10 datasets. All initial values were fixed at $10^{-5}$.
	}\label{fig:learning}
\end{figure}

We calculated the mean squared error (MSE) of posterior mean and weights on a grid of different numbers of trials and neurons.
Figure~\ref{fig:converge} shows the convergence in MSE trend. The posterior mean of the latent distribution converges to the true latent as the number of neurons grows, and the parameters converge to the true weights as the number of time bins grows.
The difference between FA and near truth initialization shows relative error in FA initialization combined with the non-convex vLGP inference which is small for latent process estimation.

We also fit the vLGP model to simulated spike trains driven by 1-dimensional Gaussian process latent at four different timescales while the rest setting was the same as above simulation. For each timescale, we simulated 10 datasets. We initialized the timescale at very smooth value ($10^{-5}$), and parameters and latent variables by factor analysis. Figure~\ref{fig:hyper} shows the learned values scattered around the ground truth for a wide range of true time scales.
We note that learning the timescale in general is challenging especially in high-dimensional latent spaces (data not shown).

\subsection{Evaluation}
We use a leave-one-neuron-out prediction likelihood to compare models.
For each dataset comprising of several trials, we choose one of the trials as test trial and the others as training trials.
First, the weights and posterior are inferred from the training trials.
Next, we leave one neuron out of the test trial and make inference on the posterior using the remaining neurons with the weights estimated from the training trials.
Then the spike train of the left-out neuron is predicted by the model given the weights estimated from the training trials and the posterior inferred from the test trial.
We repeat this procedure on each neuron of the chosen test trial, and choose each trial of one dataset as test trial.
Finally we obtain the prediction of all spike trains in the dataset.

% rank correlation
For simulated datasets, we know the true latent process that generates observations.
Since latent space is only identifiable up to affine transformation, we can quantify using the angle between subspaces~\cite{Buesing2012, Pfau2013}.
However, due to possible mismatch in the point nonlinearity, the subspace can be distorted.
To account for this mismatch, we use the mean Spearman's rank correlation that allows invertible monotonic mapping in each direction.
The Spearman's rank correlation between the posterior and the true latent trajectory gives a measure of the goodness of the posterior.
If the correlation is large, the posterior recovers more information about the underlying trajectory.
%For each sample, we compute the correlation between the posterior and true latent after concatenate all the trials along time.
%Since the models do not assume continuity between trials and impose constrains on the posterior mean, the concatenation is performed before computing the correlation.

\begin{figure}[htb!]
	\centering
	\begin{subfigure}{\textwidth}
		\centering
		\includegraphics[width=\textwidth]{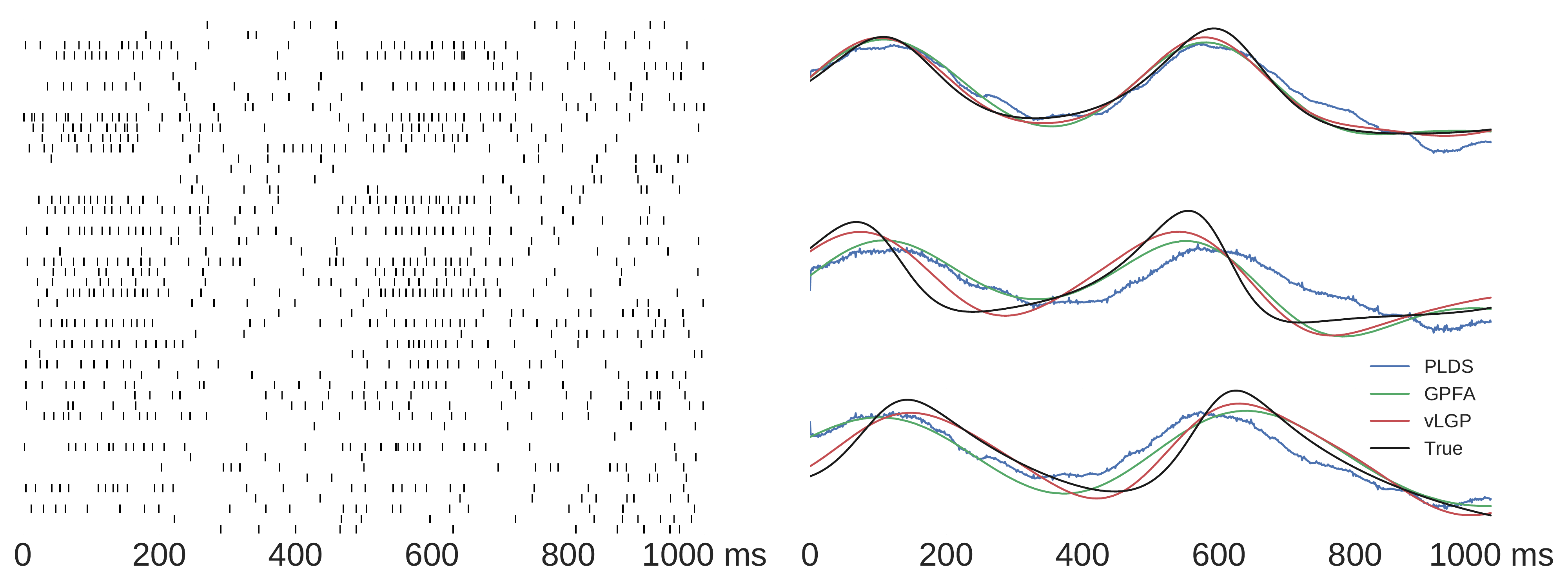}
		\caption{Lorenz attractor with refractory period}\label{fig:spk_lorenz}
	\end{subfigure}
	\\
	\begin{subfigure}{\textwidth}
		\centering
		\includegraphics[width=\textwidth]{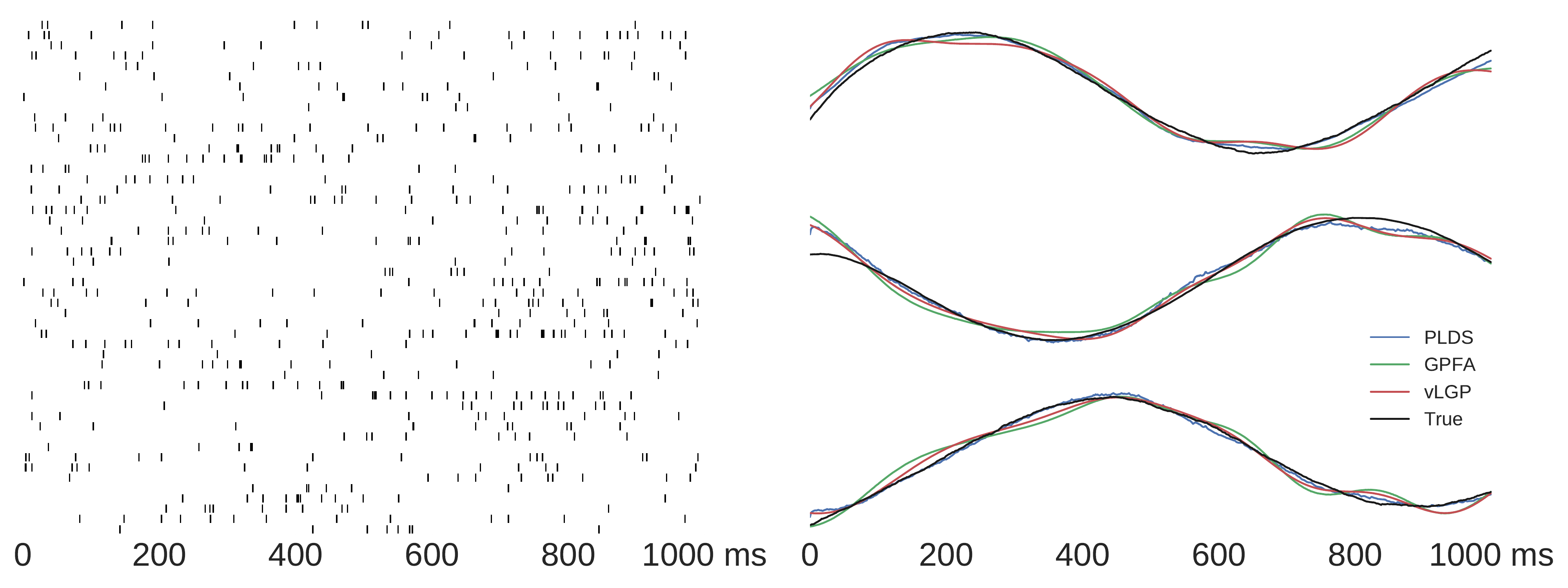}
		\caption{Linear dynamical system (LDS) with soft-rectified Poisson observation.}\label{fig:spk_plds}
	\end{subfigure}
	\caption{
		Spike trains from $50$ simultaneously observed neurons, and corresponding $3$-dimensional latent dynamics.
		\textbf{(Left)} Simulated spike trains from each corresponding system.
		See \eqref{eq:lorenz} and \eqref{eq:plds} for the exact generative model.
		\textbf{(Right)} True and inferred $3$-dimensional latent processes.
		vLGP and GPFA infers smooth posterior, while noticeable high-frequency noise is present in the PLDS inference.
	}\label{fig:spike_and_latent}
\end{figure}

\subsection{Simulation}
We simulate two datasets: one with deterministic nonlinear dynamics, and one with linear dynamics and model-mismatched nonlinear observation.
Each dataset consists of $5$ samples (simulated datasets) and each sample contains $10$ trials from $50$ neurons which last for $1$ sec. We choose a bin size of $1$~ms. 

In the first dataset, the latent trajectories are sampled from the Lorenz dynamical system with the time step of $0.0015$.
This 3-dimensional system is defined by the following set of equations,
\begin{equation}\label{eq:lorenz}
\begin{split}
\dot{x} &= 10 (y - x), \\
\dot{y} &= x (28 - z) - y, \\
\dot{z} &= x y - 2.667 z.
\end{split}
\end{equation}
Spike trains are simulated by \eqref{eq:intensity} with $10$-step suppressive history filter (from most recent: $[-10, -10, -3, -3, -3, -3, -2, -2, -1, -1]$) given the latent trajectory.
% discrete version
%\begin{equation}\label{eq:lorenz}
%\begin{split}
% x_t &= x_{t-1} + 10(y_{t-1} - x_{t-1}) \Delta, \\
% y_t &= y_{t-1} + [x_{t-1}(28 - z_{t-1}) - y_{t-1}] \Delta, \\
% z_t &= z_{t-1} + [x_{t-1} y_{t-1} - 2.667 z_{t-1}] \Delta.
%\end{split}
%\end{equation}

In the second dataset, Poisson spike trains are simulated from a $3$-dimensional linear dynamical system (LDS) defined as
\begin{equation}\label{eq:plds}
\begin{split}
y_{t,n} \given \bm{x}_t      &\sim \mathrm{Poisson}(\log(1+\exp(\bm{c}_n^\tp \bm{x}_t + \bm{d}_n)) \\
\bm{x}_0                     &\sim \mathcal{N}(\bm{\mu}_0, \bm{Q}_0) \\
\bm{x}_{t+1} \given \bm{x}_t &\sim \mathcal{N}(\bm{A} \bm{x}_t + \bm{b}_t, \bm{Q}).
\end{split}
\end{equation}

Figure~\ref{fig:spike_and_latent} shows one trial from each dataset and corresponding inferred posterior mean latent process.
The posterior means are rotated toward the true latent subspace. 
The PLDS inference (blue) looks the farthest away from the true Lorenz latent relatively but much closer to the LDS latent because the true latent meets its assumption. However, PLDS inferred latents lack smoothness.
The GPFA inference (green) is better than PLDS for Lorenz latent but shows deviations from the true LDS latent. The smoothness is kept in the inference.
The inference of our method (red) are very close to the true latent in both cases along the time while being smooth at the same time.

\begin{figure}
	\centering
	\begin{subfigure}{0.49\textwidth}
		\centering
		\includegraphics[width=\textwidth]{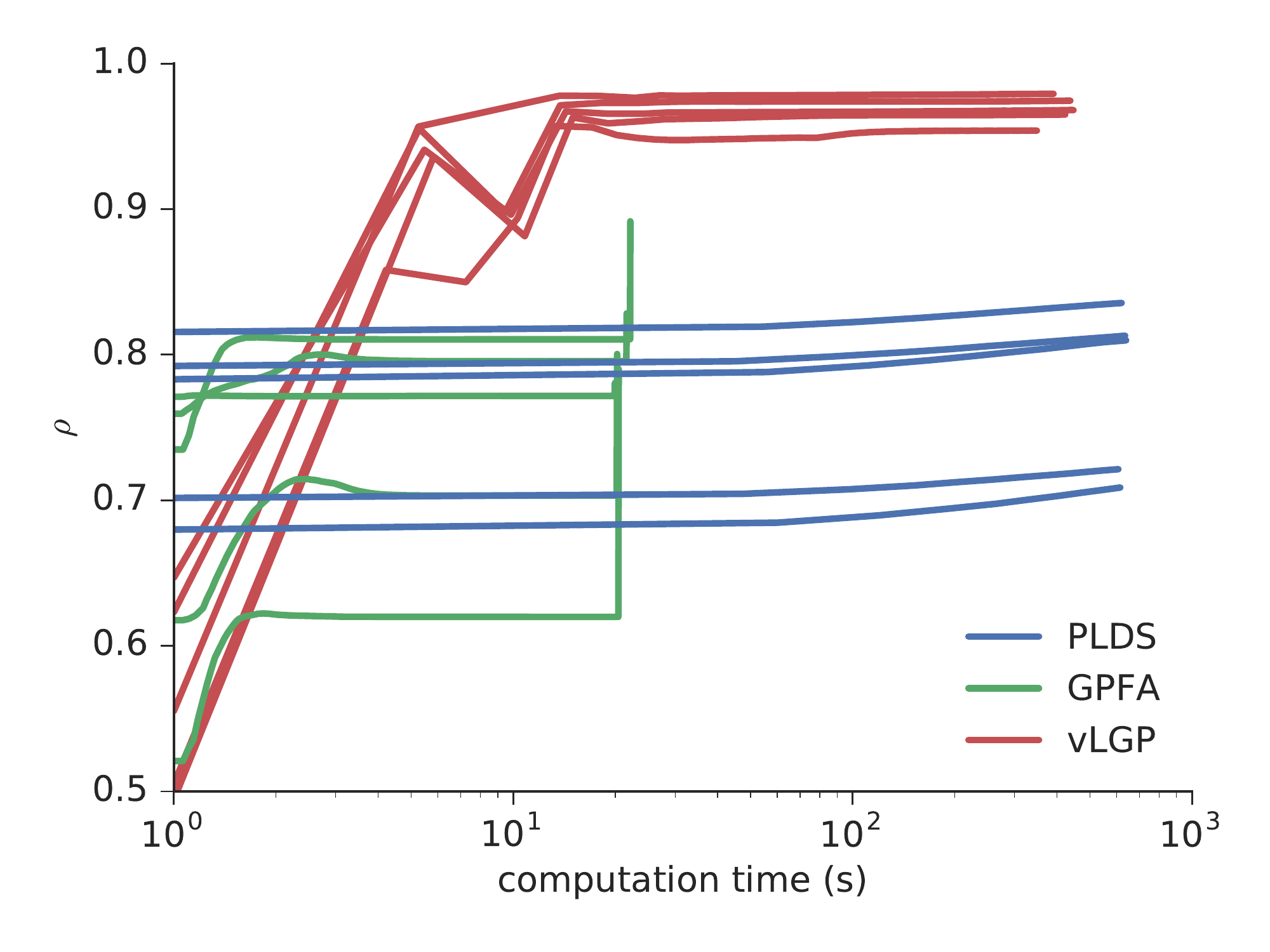}
		\caption{Lorenz}\label{fig:lorenz_corr}
	\end{subfigure}
	\begin{subfigure}{0.49\textwidth}
		\centering
		\includegraphics[width=\textwidth]{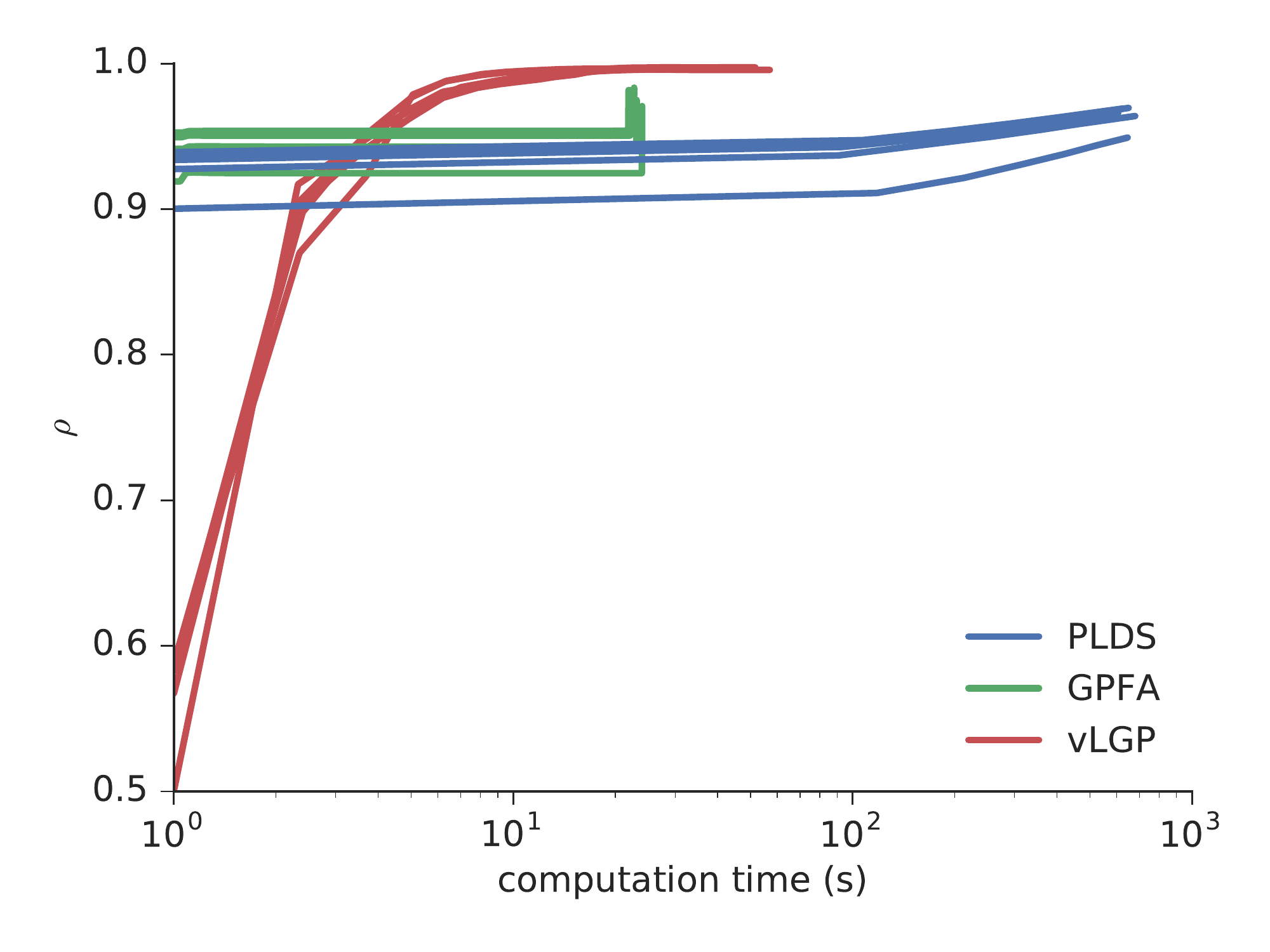}
		\caption{LDS}\label{fig:lds_corr}
	\end{subfigure}
	\caption{
		Performance comparison on simulated datasets.
		\textbf{(a,b)} Convergence speed of each algorithm in terms of inferred rank correlation between the true generative latent time series and the inferred mean posterior.
		GPFA is the fastest, and PLDS converges very slowly.
		vLGP achieves the largest correlation, yet an order of magnitude faster than PLDS.
		The origin of time is shifted to $1$ for convenience.
		Computer specification: Intel(R) Xeon(R) CPU E5-2680 v3 2.50GHz, 24 cores, 126GB RAM.
	}\label{fig:corr_of_latent}
\end{figure}

Figure~\ref{fig:corr_of_latent} shows the Spearman's rank correlation between the posterior mean and true latent versus running time (log scale).
The figures show our method (vLGP) resulted in overall larger correlation than the PLDS and GPFA after the algorithms terminated.
PLDS uses nuclear norm penalized rate estimation as initialization~\cite{Pfau2013}.
The rank correlation from PLDS inference only slowly improved from the initial value through the optimization.
Both the GPFA and vLGP use factor analysis as initialization~\cite{Yu2009}.
Note that the GPFA divides each trial into small time segments for estimating the loading matrix and bias which it breaks the continuity within each trial.
Only the final iteration infers each trial as whole.
Thus the correlations of the final iterations jumps up in the figures.
It is obvious that vLGP makes much improvement to the result of factor analysis in terms of the rank correlation. 

To quantify predictive performance on the spike trains, we use the log-likelihood on the leave-one-neuron-out as described in the evaluations section.
We normalize the test point process likelihood with respect to that of a baseline model that assumes a homogeneous Poisson process to obtain, the predictive log-likelihood (PLL), given as,
\begin{equation}
\mathrm{PLL} = \frac{\left[\sum_{t,n} (y_{y,n} \log(\lambda_{t,(-n)}) - \lambda_{t,(-n)})\right] - \left[\sum_{t,n} (y_{y,n} \log(\bar{y}) - \bar{y})\right]}{\text{(\# of spikes)}\log(2)} ,
\end{equation}
where $\lambda_{t,(-n)}$ is the leave-neuron-out prediction to the firing rate of neuron $n$ at time $t$, and $\bar{y}$ is the population mean firing rate.
Positive PLL implies the model predicts better than mean firing rate, and higher PLL implies better prediction.
PLL has a unit of \textit{bits per spike}, and is widely used to quantify spike train prediction~\cite{Pillow08}.

\begin{figure}
	\centering
	\begin{subfigure}{0.49\textwidth}
		\centering
		\includegraphics[width=\textwidth]{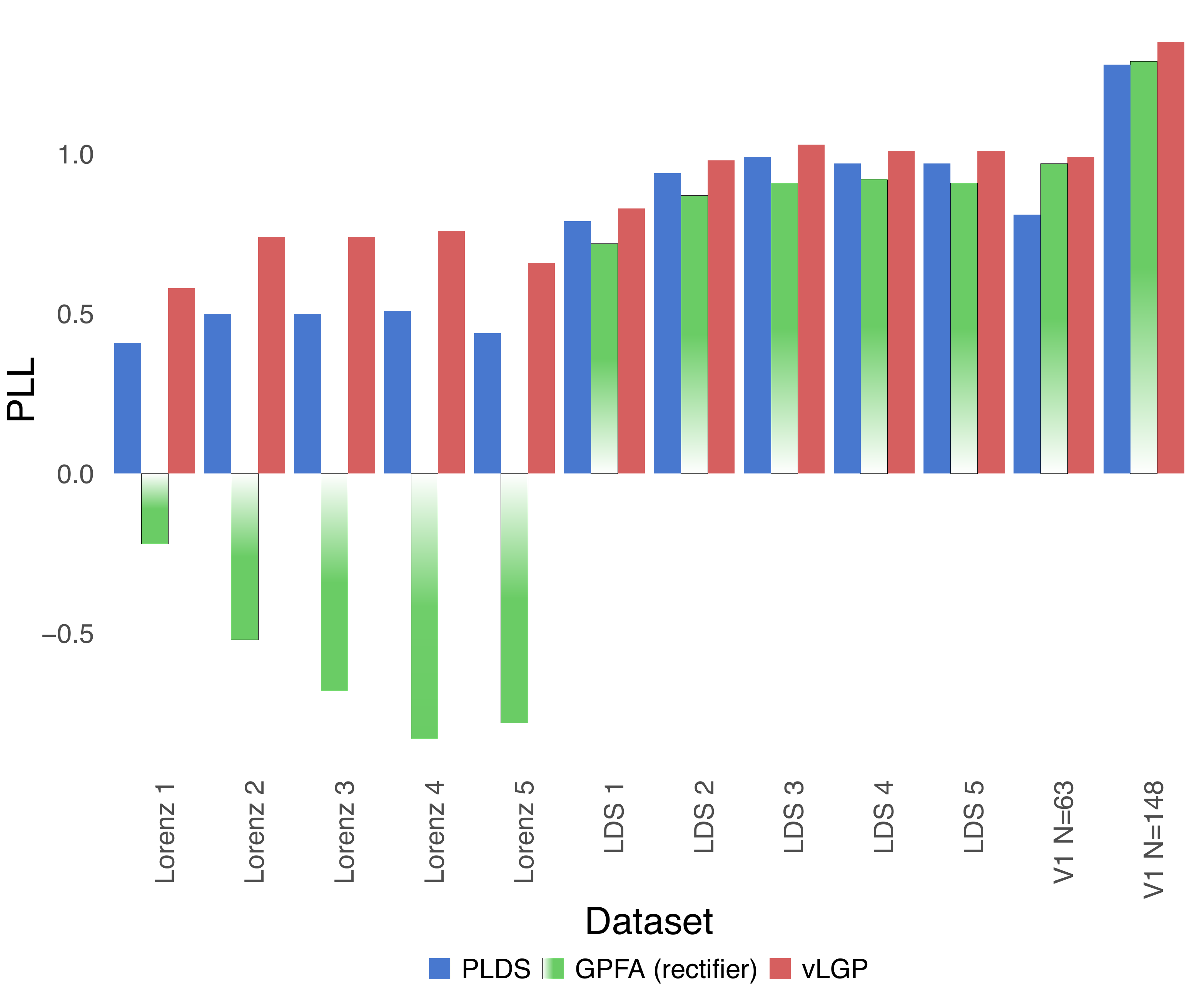}
		\caption{PLL}\label{fig:pll}
	\end{subfigure}
	\begin{subfigure}{0.49\textwidth}
		\centering
		\includegraphics[width=\textwidth]{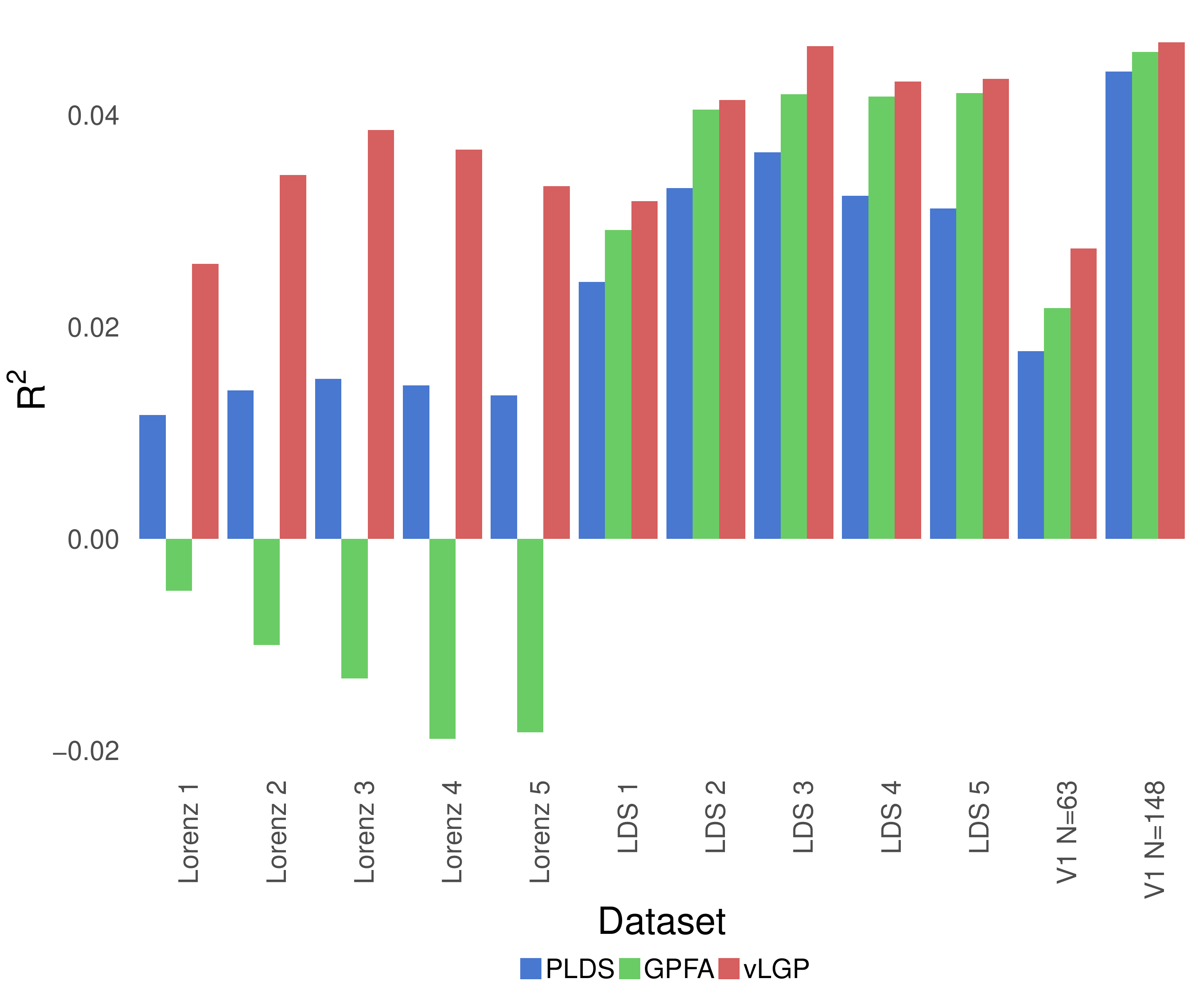}
		\caption{$R^2$}\label{fig:rsquared}
	\end{subfigure}
	\caption{
		Predictive performance (the larger the better).
		(a) Predictive log-likelihood (PLL)
		(b) Predictive $R^2$ (coefficient of determination)
		Note that GPFA can predict negative mean, thus the PLL cannot make a fair comparison even with a soft rectifier link.
		We visually added gradient in the GPFA PLL results to caution the readers.
		$R^2$, which is GPFA's natural measure of performance can also be negative due to overfitting.
		Note that GPFA performs better than PLDS on LDS when compared in predictive $R^2$, but vLGP is consistently the best in both measures.
	}\label{fig:prediction}
\end{figure}

In Figure~\ref{fig:prediction}, we compare the three models for each dataset.
Since GPFA assumes a Gaussian likelihood, it is incompatible to compare directly using a point process likelihood.
We use linear rectifier to convert the GPFA predictions to non-negative rates, then compute PLL (Fig.~\ref{fig:pll}).
Let us denote the linear predictor by $\eta$ omitting the neuron, time and model.
Specifically, The rate prediction is given by,
\begin{equation}
\lambda = 
\begin{cases}
\log(1 + \exp(a\eta)) / a                            & \text{GPFA (rectifier link)} \\
\exp(\eta + \frac{1}{2} \bm{\alpha}^\tp \bm{V} \bm{\alpha}) & \text{PLDS and vLGP}
\end{cases}
\end{equation}
where $a=500$ was chosen to prevent the prediction from producing invalid PLL while preserving linearity as much as possible\footnote{We tried square link function for GPFA initially. However, it often produces detrimental PLL due to large rate predictions from large negative raw ($\eta$) predictions. Also note that straightforward linear rectifier $\lambda = \eta \cdot \mathbb{I}(\eta > 0)$ can result in undefined PLL due to predicting zero rate in a bin with non-zero observation.}.
We also compare the $R^2$ instead of PLL of the predictions with respect to the three models (Fig.~\ref{fig:rsquared}) where the raw linear prediction $\eta$ without the rectifier link was used particularly.
PLL and $R^2$ largely agrees despite that $R^2$ assumes a squared error measure GPFA optimizes for, and inappropriate for PLDS and vLGP.

\subsection{V1 population recording} % Graf dataset
We apply our method to a large scale recording to validate that vLGP picks up meaningful known signals, and investigate the population-wide trial-to-trial variability structure.
We use the dataset~\cite{Graf2011} where 72 different equally spaced directional drifting gratings were presented to an anesthetized monkey for 50 trials each (array-5, 148 simultaneously recorded single units). 
We use 63 V1 neurons by only considering neurons with tuning curves that could be well approximated ($R^2 \ge 0.75$) by bimodal circular Gaussian functions (the sum of two von Mises functions with different preferred orientations, amplitudes and bandwidths) according to~\cite{Graf2011}.
We do not include the stimulus drive in the model, in hopes that the inferred latent processes would encode the stimulus.
We used bin size of 1~ms.
%The repeated fixed stimulus presentations allow us to verify that the inferred latent processes indeed encode the stimulus.

%The spike trains were recorded from 148 V1 neurons for 2560~ms. The time bin is 1~ms. The first 1280~ms is stimulus and the second 1280~ms is blank.
% The stimulus have 72 directions. There were 50 trials performed for each direction. We combine the first 5 trials at 0 degree and the first 5 trials at 90 degree as one sample of 10 trials. 

%\begin{figure}[th!]
%   \centering
%   \includegraphics[width=\textwidth]{graf/graf5_63n_LL_vs_Dim}
%   \caption{Log-likelihood vs Latent Dimension}
%   \label{fig:graf5_63n_LL_vs_Dim}
%\end{figure}

We use 4-fold cross-validation to determine the number of latents. A 15--dimensional model is fitted to a subsample composed of the first trial of each direction at first. In each fold, we use its estimated parameter to infer the latent process from another subsample composed of the second trial of each direction. The inference is made by leaving a quarter of neurons out, and we predict the spike trains of the left-out neurons given the first $k$ ($k=1 \ldots 15$) orthogonalized latent process corresponding to $k$-dimension.
This procedure led us to choose 5 as the dimension since the predictive log-likelihood reached its maximum.

We re-fit a 5-dimensional vLGP model using the subsample of the first trials. To quantify how much the model explains, we report pseudo-$R^2$ defined as
\begin{equation}
\text{pseudo-$R^2$} = 1 - \frac{\mathrm{LL_{saturated}} - \mathrm{LL_{model}}}{\mathrm{LL_{saturated}} - \mathrm{LL_{null}}}
\end{equation}
where $\mathrm{LL_{null}}$ refers to the log-likelihood of population mean firing rate model (single parameter), and $\mathrm{LL_{saturated}}$ is the log-likelihood for the saturated model in which the rate of each bin is estimated by the empirical mean. 
The pseudo-$R^2$ of our model (vLGP with 5D latents) is 20.88\%. This model explains with
shared variability through the latents, and heterogeneity of baseline firing of individual neurons. For a
baseline model with only per neuron noise component (and no shared latent), the pseudo-$R^2$ is 6.77\%. 

Figure~\ref{fig:prediction} shows the predictive performance based on two subsets. The first one is 4 trials (0$^\circ$, 90$^\circ$, 180$^\circ$, 270$^\circ$) of the subset of 63 neurons with 5-dimensional latent process. The second one is 10 trials (5 trials of 0$^\circ$ and 5 trials of 90$^\circ$) of all 148 neurons with 4-dimensional latent process. 

To evaluate the predictive performance under a longer time scale, we also cross-validated on the first subset with 20~ms time bins with GPFA and vLGP. GPFA models were fitted for both raw spike count and its square root. The mean of square root spike count by vLGP was obtained from simulation using the predicted firing rate. We report the normalized MSEs (MSE / variance of observation) of both methods for spike count and its square root respectively, GPFA: 0.713 (spike count) and 0.699 (square root), and vLGP: 0.709 (spike count) and 0.708 (square root). We use F-test to compare the MSEs and see if any one of the two methods results in a significantly larger error. The corresponding p-values are 0.588 (spike count) and 0.140 (square root). It shows that the MSEs are not significantly different between GPFA and vLGP with 20~ms time bin.

%Figure~\ref{fig:graf_elbo} shows the ELBOs up to $9$ latents.
%The maximum ELBO appears at $4$ latents.
%Therefore we use four latent vLGP model for subsequent analyses.

%\begin{figure}
%   \centering
%   \includegraphics[width=0.5\textwidth]{graf_elbo}
%   \caption{Model selection on latent dimensionality for vLGP using all trials from 0 and 90 degrees directional drifting grating. Note that ELBO peaks at four latents, hence we choose a four dimensional model.
%}\label{fig:graf_elbo}
%\end{figure}

\begin{figure}[tbh]
	\centering
	\begin{subfigure}[t]{0.49\textwidth}
		\includegraphics[width=\textwidth]{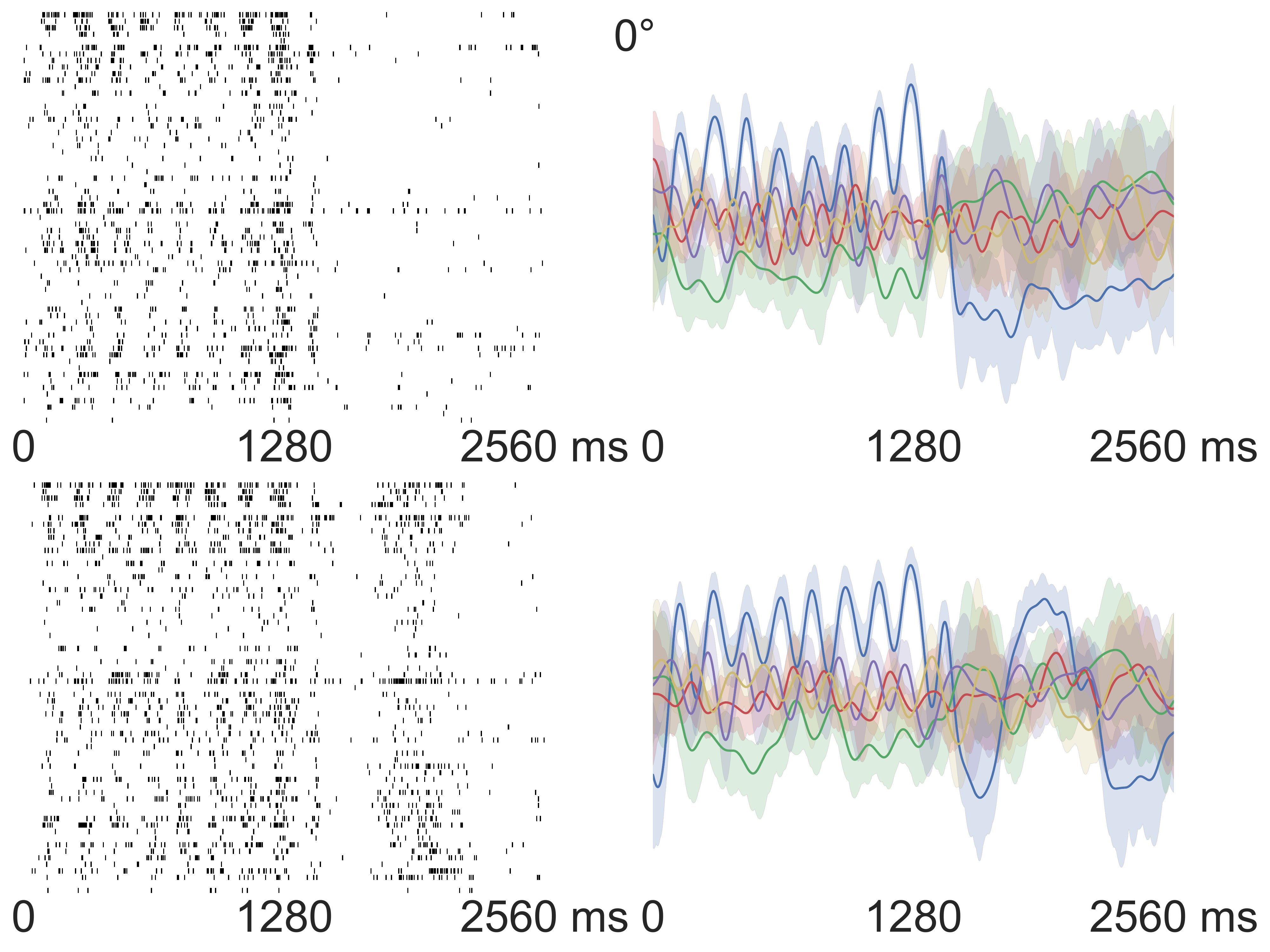}
	\end{subfigure}
	\begin{subfigure}[t]{0.49\textwidth}
		\includegraphics[width=\textwidth]{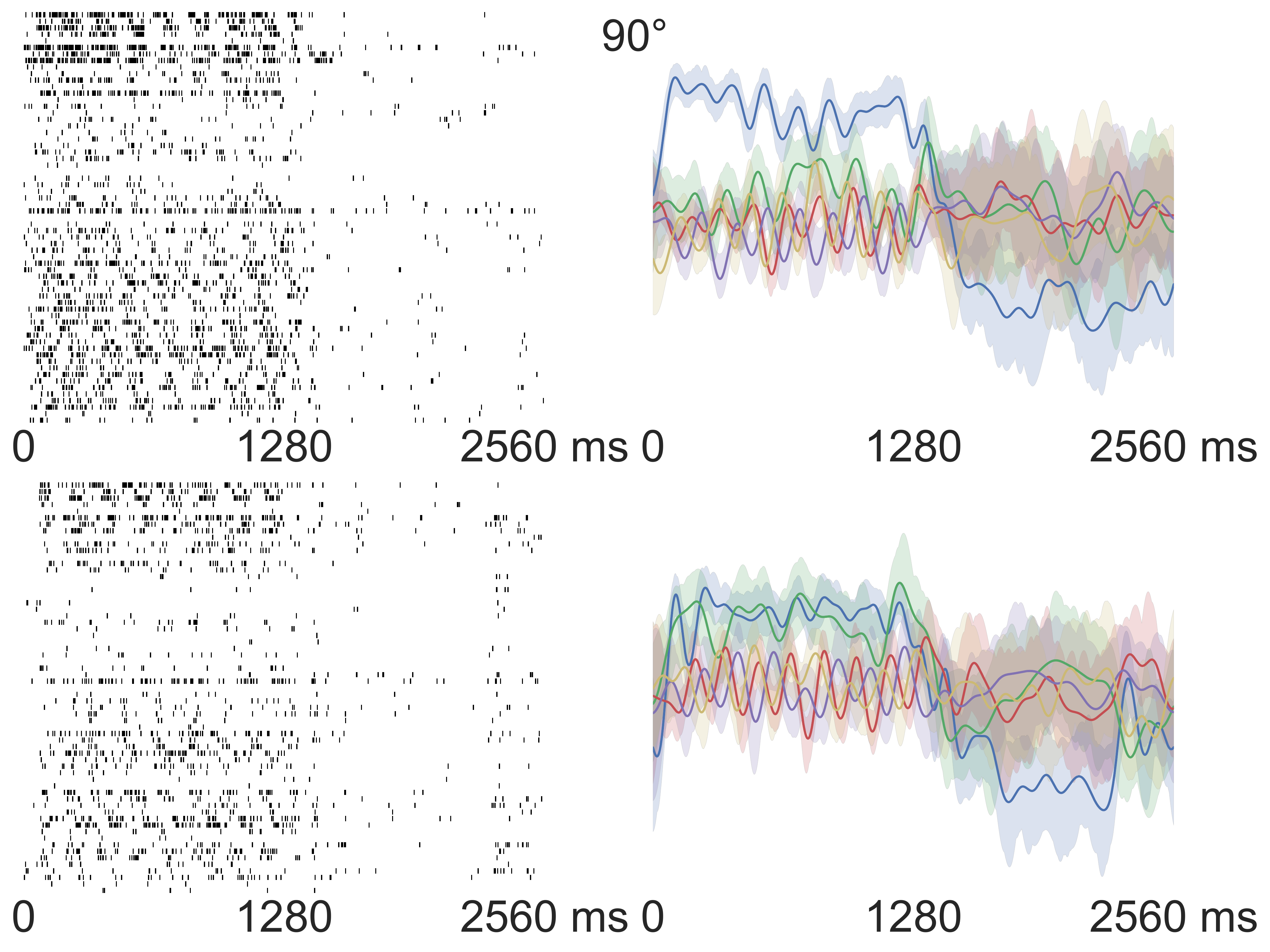}
	\end{subfigure}
	\caption{Single trial spike trains and inferred latent.
		The visual stimulus was only on for the first half of the trial.
		The left two columns are the spike trains and respective inferred latent of 2 trials of $0^{\circ}$.
		The right ones are 2 trials of $90^{\circ}$.
		The colors indicate the latent dimensions that are rotated to maximize the power captured by each latent in decreasing order (\textcolor{blue}{blue}, \textcolor{green}{green}, \textcolor{red}{red}, \textcolor{purple}{purple} and \textcolor{yellow}{yellow}). The solid lines are the posterior means and the light colors are corresponding uncertainty.
	}
	\label{fig:latent:singletrial}
\end{figure}

Although the parameters are estimated from a subsample, we can use them to infer the latent process of all trials of all 72 directions.
Figure~\ref{fig:latent:singletrial} shows inferred latent processes for two trials for two directions. We rotate the inferred latent process by the singular value decomposition (SVD; details will be given later.)
Variational posterior distribution over the latents are shown for each trial.
During the second half of the trial when the stimulus was off, and the firing rate was lower, the uncertainty in the latent processes increases.
There are visible trial-to-trial variability in the spike trains which are reflected in the variations of latents.

\begin{figure}
	\hfill
	\centering
	\begin{subfigure}[t]{0.49\textwidth}
		\includegraphics[width=\textwidth]{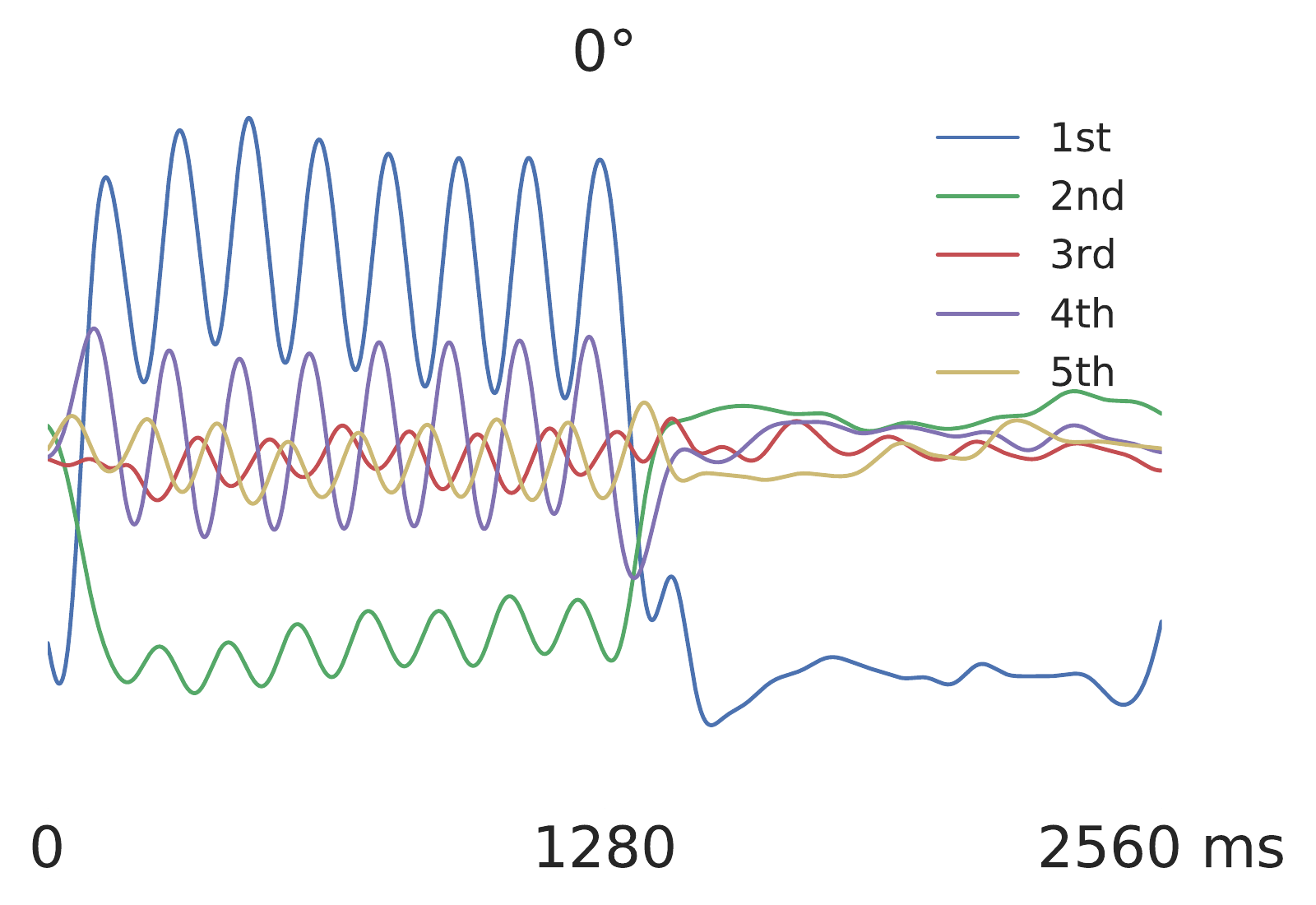}
	\end{subfigure}
	\begin{subfigure}[t]{0.49\textwidth}
		\includegraphics[width=\textwidth]{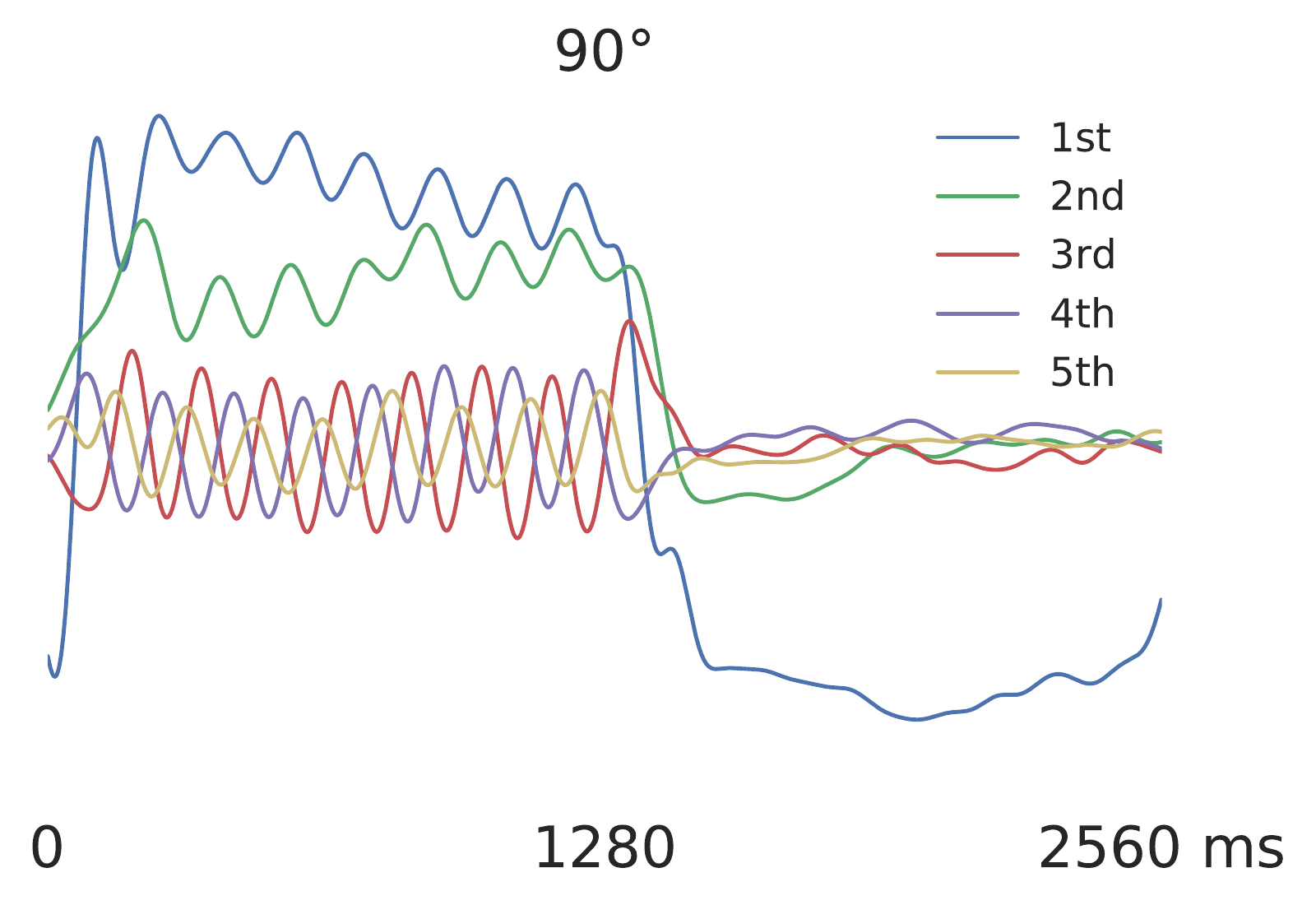}
	\end{subfigure}
	\caption{
		Inferred latent processes averaged for two stimulus directions ($0\si{\degree}$ and $90\si{\degree}$). 
		Latents are rotated to maximize the power captured by each latent in decreasing order.
	}
	\label{fig:latent:mean}
\end{figure}

First we investigate how the ``signal''---defined as visual stimuli---is captured by the latent processes.
We average the inferred latent processes over 50 trials with identical spatio-temporal stimuli (Fig.~\ref{fig:latent:mean}).
Since the stimuli are time-locked, corresponding average latent trajectory should reveal the time-locked population fluctuations driven by the visual input. We concatenate the average latent processes along the dimension of time. Then we orthogonalize it by SVD. The dimensions of orthogonalized one are ordered by the respective singular values. The latent process of a single trial is also rotated to the same subspace. 

\begin{figure}
	\centering
	\begin{subfigure}[t]{0.49\textwidth}
		\includegraphics[width=\textwidth]{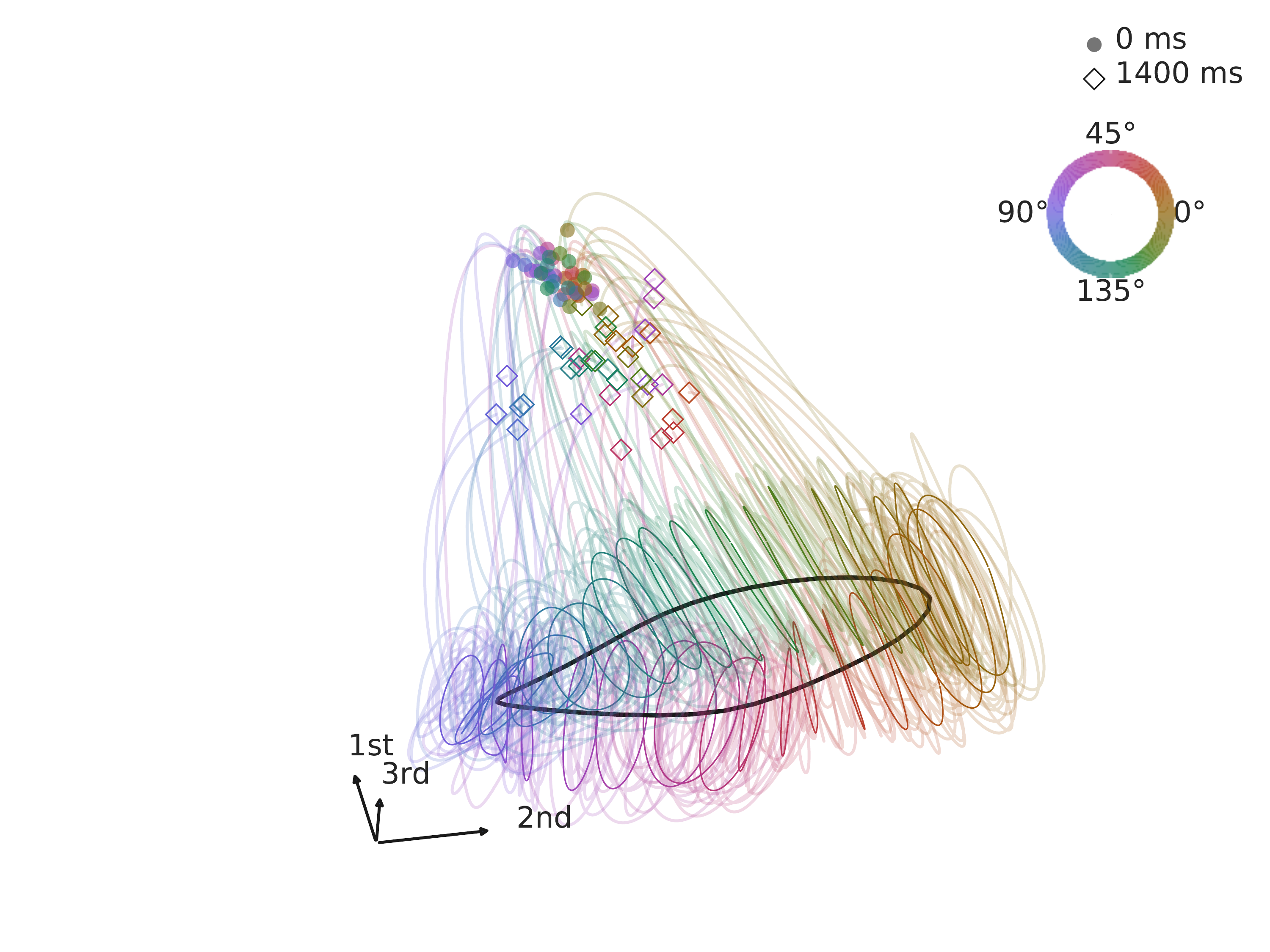}
	\end{subfigure}
	\begin{subfigure}[t]{0.49\textwidth}
		\includegraphics[width=\textwidth]{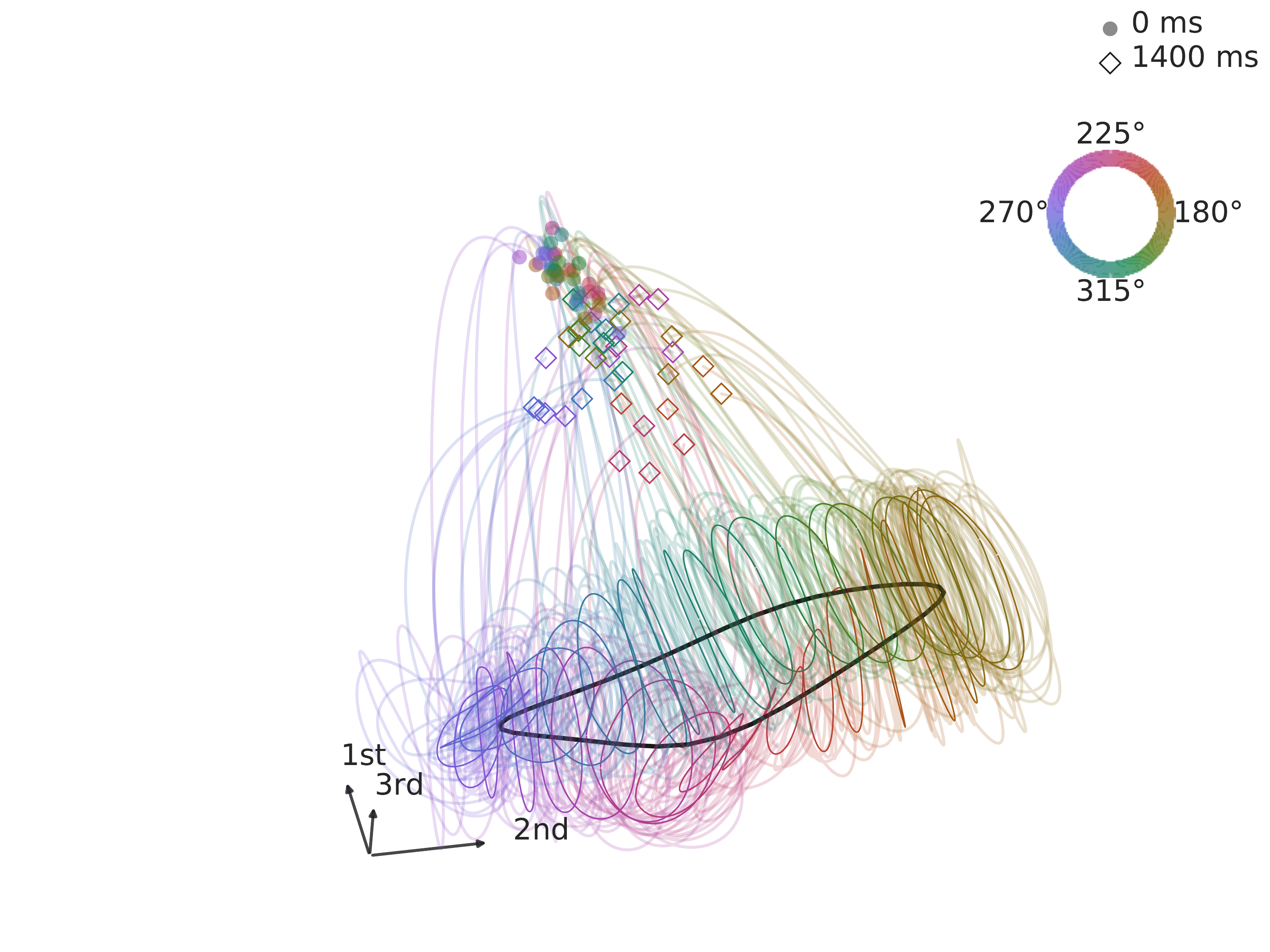}
	\end{subfigure}  
	\caption{3D projection of mean latent trajectories given each orientation.
		We plot the first three singular vectors of the inferred latent corresponding to the signal interval (0--1400~ms) colored by orientation.
		The colored circles are cycle averages that visualize the temporal phase of oscillation per direction, and form an approximate torus.
		The black circle visualizes the circular orientation that goes through the center of the torus.
		The left side shows 0--180$^{\circ}$ and the right side shows 180--360$^{\circ}$.
	}
	\label{fig:latent:3D_proj}
\end{figure}

Furthermore, we visualized the trajectories in 3D (see supplementary online video\footnote{\url{https://www.youtube.com/watch?v=CrY5AfNH1ik}}) that show how signal and noise are dynamically encoded in the state space.
Figure~\ref{fig:latent:3D_proj} shows the projection of average latent process corresponding to each orientation to the first 3 principal components.
The projection topologically preserves the orientation tuning in the V1 population.
There are two continuous circular variables in the stimuli space to be encoded: orientation and temporal phase of oscillation.
The simplest topological structure of neural encoding is a torus, and we observe a toroidal topology (highlighted as rings of cycle averages).

%We also did cross-validation on the dataset that consists of 5 trials of $0^\circ$ and 5 trials of $90^\circ$. We report the PLL w.r.t. three models, $\mathrm{PLL}_\mathrm{PLDS} = \SI[mode=math]{1.28}{bit \per spike}$, $\mathrm{PLL}_\mathrm{GPFA} =  \SI[mode=math]{-2.45}{bit \per spike}$ and $\mathrm{PLL}_\mathrm{vLGP} =  \SI[mode=math]{1.35}{bit \per spike}$.

\begin{figure}
	\centering
	\includegraphics[width=\textwidth]{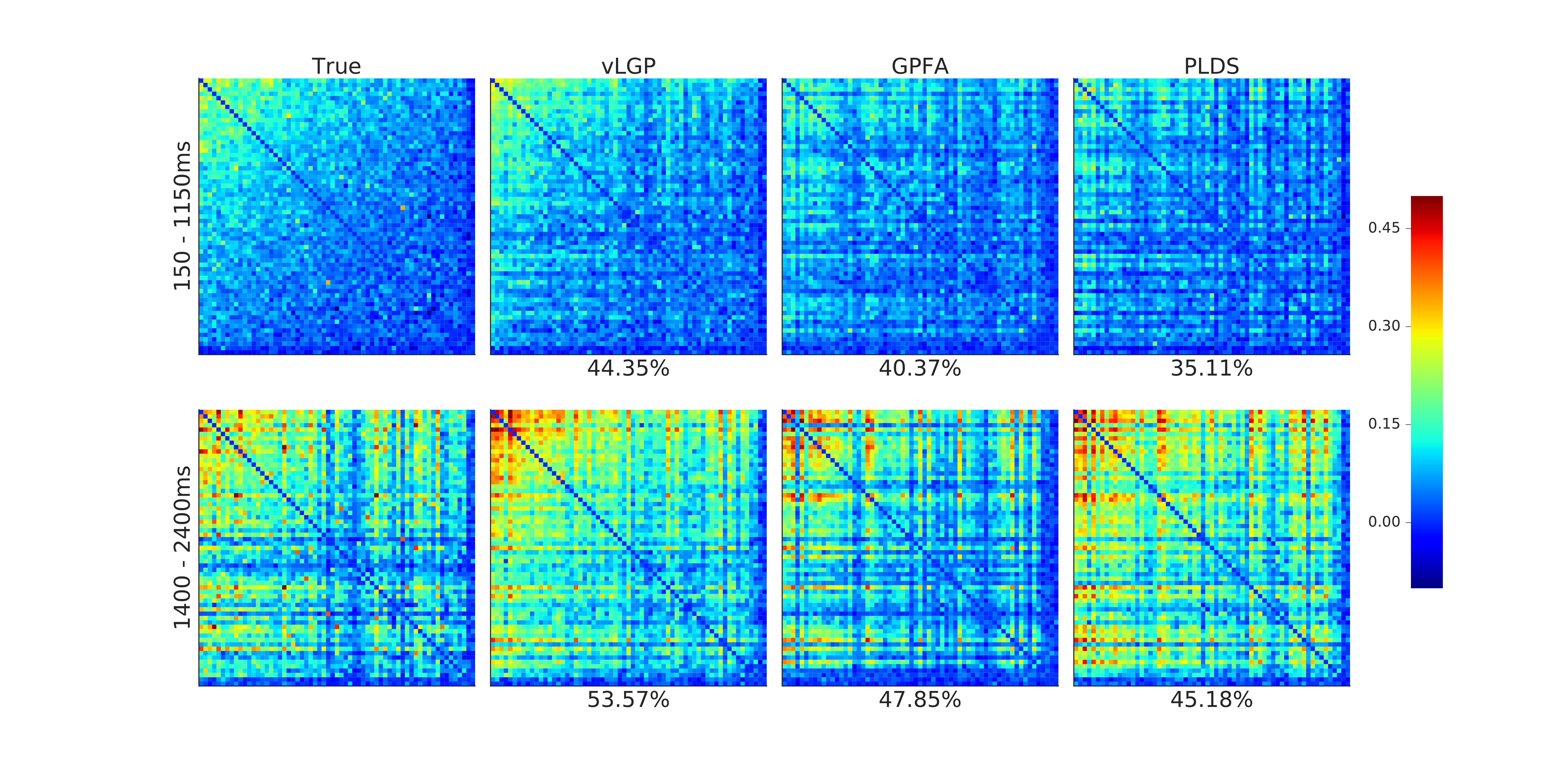}
	\caption[Noise-correlation]{Noise-correlation analysis.
		The pairwise noise correlations between all neurons were calculated during the stimulus period (top, 150--1150~ms) and off-stimulus period (bottom, 1400--2400~ms).
		The time bin size is 50~ms.
		Neurons are sorted by the total noise-correlation defined as the row sum of stimulus-driven noise correlation matrix (top-left).
		The model-explained power percentages are shown on the bottom of each matrix.
	}\label{fig:graf5_corr}
\end{figure}

To see if our method captures the noise correlation structure through the latents as one would predict from recent studies of cortical population activity~\cite{Goris2014,Ecker2014b,Lin2015}, we calculated pairwise correlations between all neurons.
We simulated spike trains by model-predicted firing rates of all trials with $0^\circ$ and $90^\circ$ stimulus. The model-predicted firing rates are the fitted values of the time bins that are calculated in the way of the generative model by using estimated parameters and latents. For each bin, we simulated the spike by using a Poisson random number generator with the corresponding firing rate.  
To remove the signal, we subtracted the mean over 50 trials for each direction of stimulus.
Figure~\ref{fig:graf5_corr} shows the correlation matrices during the stimulus period (150--1150~ms) and off-stimulus period (1400--2400~ms).
The neurons are sorted by the total correlations during the stimulus period.
The power of model-explained noise correlation is defined as
$(1 - {\lVert \bm{C}_{\mathrm{model}} - \bm{C}_{\mathrm{true}}\rVert_{\mathrm{F}}})/{\lVert \bm{C}_{\mathrm{true}}\rVert_{\mathrm{F}}}$
where $\bm{C}$ is the zero-diagonal correlation matrix w.r.t. its subscript and $\lVert \cdot \rVert_{\mathrm{F}}$ is the Frobenius norm.
The proposed model explains more noise correlation in contrast to GPFA and PLDS for both periods.

These results show that vLGP is capable of capturing both the signal---repeated over multiple trials---and noise---population fluctuation not time locked to other task variables--present in the cortical spike trains.

\section{Discussion}
We propose vLGP, a method that recovers low-dimensional latent dynamics from high-dimensional time series.
Latent state-space inference methods are different from methods that only recover the average neural response time-locked to an external observation~\cite{Churchland2012,Brendel2011}.
By inferring latent trajectories on each trial, they provide a flexible framework for studying the internal neural processes that are not time-locked.
Higher-order processes such as decision-making, attention, and memory recall are well suited for latent trajectory analysis due to their intrinsic low-dimensionality of computation.
Our method performs dimensionality reduction on a single trial basis and allows decomposition of neural signals into a small number of smooth temporal signals and their relative contribution to the population signal.

We compare our method to two widely used latent state space modeling tools in neuroscience: GPFA~\cite{Yu2009} and PLDS~\cite{Macke2011c}.
Unlike GPFA, vLGP allows a generalized linear model observation which is more suitable for a broad range of spike train observations.
Moreover, vLGP is significantly faster than PLDS, yet it shows superior performance in capturing the spatio-temporal structures in the neural data to both PLDS and GPFA at a fine timescale (1~ms bins).
% as the reviewer asked 

To test its validity in real electrophysiological recordings, we used V1 population recording driven by fixed stimulus as a litmus test.
We showed that our inferred latents contain meaningful information about the external stimuli, encoding both orientation and temporal modulation on a continuous manifold.

We only considered smoothness encoded in the GP prior in this manuscript, but a plethora of GP kernels are available~\cite{Rasmussen2005,Scholkopf2002}.
For example, to capture prior assumptions about periodicity in some of the latent processes, we can use spectral kernels~\cite{Ulrich2015}.
This can be particularly useful for capturing internal neural oscillations~\cite{Fries2001}.
In addition, it is straightforward to incorporate additional covariates such as external stimuli~\cite{Park2014d} or local field potential~\cite{Kelly2010} to vLGP.

% weakness
We have not found systematic issues with vLGP, but potential weaknesses could stem from the variational approximation, inappropriate assumptions on the latent processes, and particular numerical shortcuts used in the implementation.
It could be challenging for our method to recover very fast-changing latent trajectories, especially when the overall firing rate is very low.
Subsampling used while optimizing the hyperparameters may miss a rare but key spike when the spike trains are sparse.

The proposed method has potential application in many areas, and it will be particularly useful in discovering how specific neural computations are implemented as neural dynamics.
We are working on applying this method and its extensions to sensorimotor decision-making process where the normative model guides what is being computed, but it is unclear as to how the neural system implements it.

An open-source python implementation of vLGP is available online (\url{https://github.com/catniplab/vLGP}) under MIT license.

\section*{Acknowledgment}
We thank the reviewers for their constructive feedback.
We are grateful to Arnulf Graf, Adam Kohn, Tony Movshon, and Mehrdad Jazayeri for providing the V1 dataset.
We also thank Evan Archer, Jakob Macke, Yuanjun Gao, Chethan Pandarinath, and David Sussillo for helpful feedback.
This work was partially supported by the Thomas Hartman Foundation for Parkinson's Research.

\bibliographystyle{apalike}
\bibliography{ref}

\end{document}